\documentclass[a4paper,fleqn]{cas-dc}
\usepackage[numbers]{natbib}

\usepackage{amssymb}
\usepackage{amsmath}
\usepackage{arydshln}
\usepackage{nicematrix}
\usepackage{multirow}
\usepackage{paralist}
\usepackage{amsthm}
\usepackage{subcaption}
\usepackage{mathrsfs}
\usepackage{booktabs}
\usepackage{tikz}
\usepackage{adjustbox}
\usepackage{enumitem}
\usepackage{algorithm}
\usepackage{algpseudocode}
\usepackage{float}

\begin{document}
\let\WriteBookmarks\relax
\def\floatpagepagefraction{1}
\def\textpagefraction{.001}
\shorttitle{Universal Incremental Learning}
\shortauthors{Zhou et al.}

\title[mode = title]{Universal Incremental Learning: Mitigating Confusion from Inter- and Intra-task Distribution Randomness}

\author[1,2]{Yi Zhou}
\cormark[1]
\ead{yizhou.szcn@gmail.com}
\credit{Conceptualization, Methodology, Formal analysis, Investigation, Writing -- original draft, Writing -- review and editing, Funding acquisition, Project administration}

\author[1,2]{Jinsong Shi}
\ead{srache@seu.edu.cn}
\credit{Methodology, Software, Formal analysis, Investigation, Visualization, Writing -- review and editing}

\author[3]{Sheng Luo}
\ead{rosenluo@didiglobal.com}
\credit{Methodology, Software, Formal analysis, Investigation, Writing -- original draft}

\author[4]{Tao Zhou}
\ead{taozhou.ai@gmail.com}
\credit{Investigation, Writing -- review and editing}

\affiliation[1]{organization={School of Computer Science and Engineering, Southeast University},
                city={Nanjing}, postcode={211189}, country={China}}
\affiliation[2]{organization={Key Laboratory of New Generation Artificial Intelligence Technology and Its Interdisciplinary Applications (Southeast University)},
                city={Ministry of Education}, country={China}}
\affiliation[3]{organization={Didi Chuxing},
                city={Beijing}, postcode={100080}, country={China}}
\affiliation[4]{organization={School of Computer Science and Engineering, Nanjing University of Science and Technology},
                city={Nanjing}, postcode={210094}, country={China}}

\cortext[cor1]{Corresponding author.}

\begin{abstract}
Existing incremental learning (IL) methods rely on strong assumptions: incoming tasks either introduce exclusively new classes or new domains (i.e., Class IL and Domain IL), or increment by a fixed scale in a class- and domain-agnostic manner (i.e., Versatile IL, VIL). These assumptions severely limit their applicability in dynamic, unpredictable real-world scenarios.
In this work, we investigate Universal Incremental Learning (UIL), a more realistic setting where the model has no prior knowledge of which new classes or domains will appear across sequential tasks, nor the scale of these increments within each task.
This uncertainty hinders the model’s ability to confidently acquire knowledge across varying task distributions and to learn uniformly from imbalanced intra-task distributions.
Consequently, UIL introduces a more general IL scenario, where the model must cope with confusion stemming from both inter-task and intra-task distribution randomness.
To mitigate both sources of confusion, we propose MiCo, a simple yet effective framework for UIL.
At the inter-task level, we employ a multi-objective learning scheme to enforce accurate and deterministic predictions, further enhanced by a direction recalibration module that alleviates conflicting gradient updates.
At the intra-task level, we introduce a magnitude recalibration module to address asymmetrical optimization caused by imbalanced class distributions.
Extensive experiments on four benchmarks demonstrate that MiCo significantly outperforms existing state-of-the-art methods in both the proposed UIL and existing VIL scenarios.
\end{abstract}

%\begin{highlights}
%\item Proposes Universal Incremental Learning for random increment types and scales.
%\item Identifies dual confusion from inter- and intra-task distribution randomness.
%\item Direction recalibration and multi-objective learning resolve gradient conflicts.
%\item Magnitude recalibration mitigates optimization bias on imbalanced classes.
%\item MiCo achieves state-of-the-art performance across four diverse UIL benchmarks.
%\end{highlights}


\maketitle

\section{Introduction}
\label{sec:intro}
Incremental learning (IL) enables models to sequentially acquire new knowledge in dynamic real-world scenarios without retraining from scratch. While early approaches relied on parameter regularization and rehearsal to preserve historical representations~\cite{ewc}, recent advances increasingly leverage parameter-efficient prompt adaptation to better handle shifting distributions~\cite{dualprompt}.

\begin{figure}[]
	\centering
	\includegraphics[width=\columnwidth]{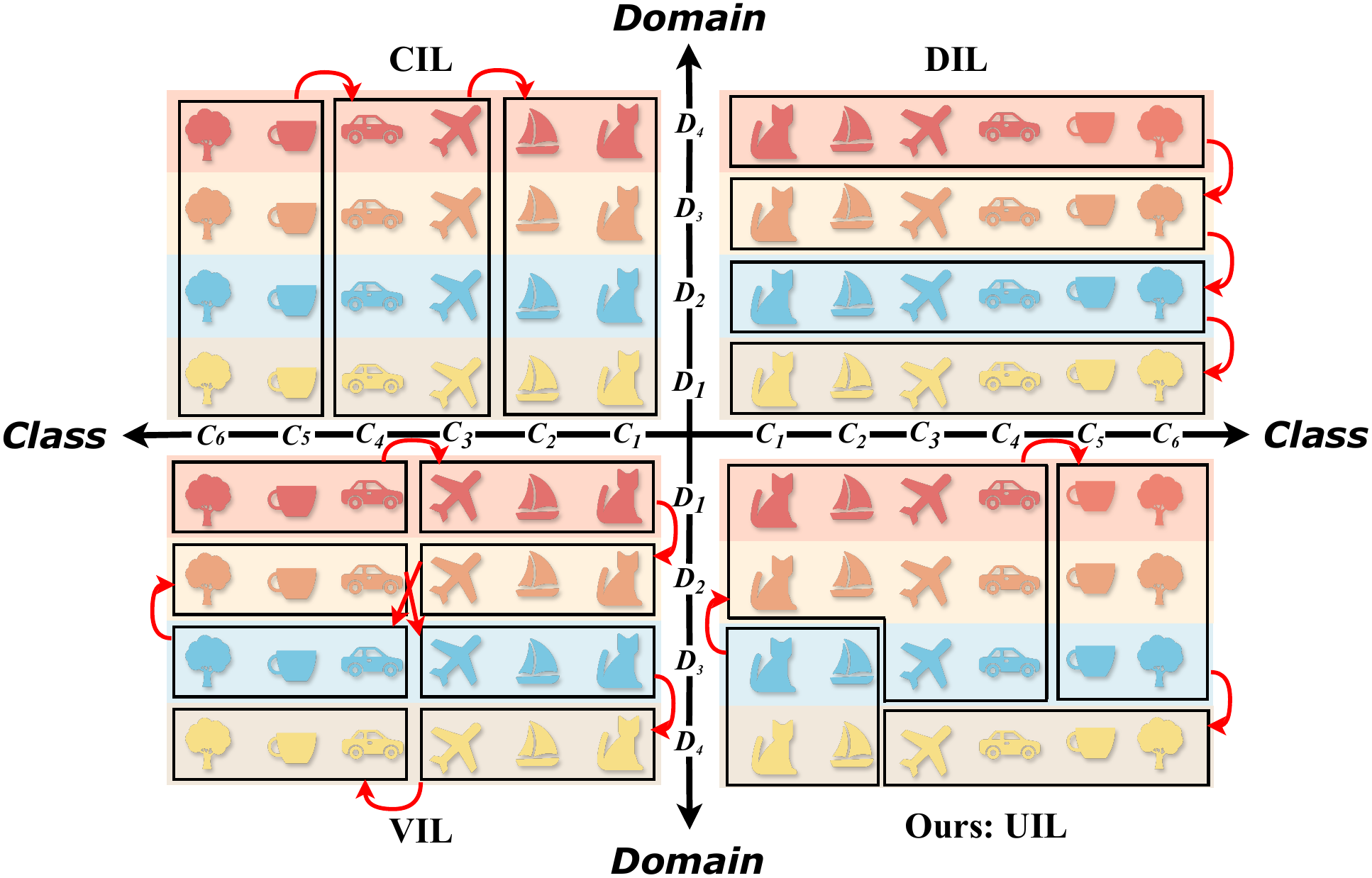}
	\caption{Comparison of existing IL scenarios and our proposed UIL scenario on a 6-class dataset with 4 domains. The horizontal and vertical axes denote class and domain indices, respectively. Each incremental task is outlined in black. CIL, DIL, and VIL assume that each incoming task introduces only class increments, domain increments, or both class and domain increments, respectively. In contrast, UIL represents a more general IL scenario by simultaneously introducing randomness in both the type and scale of increments, where the model has no prior knowledge of which new classes or domains will emerge across sequential tasks, nor of the number of new classes or domains within each task.}   
	\label{fig:scenario}
\end{figure}

Based on the nature of incoming knowledge, mainstream IL scenarios are primarily categorized into Class-IL (CIL)~\cite{l2p, chen2025make} and Domain-IL (DIL)~\cite{sprompts,selfsup}, as illustrated in Fig.~\ref{fig:scenario}. Specifically, CIL assumes that each incremental task introduces a disjoint set of new classes within the same domain, whereas DIL focuses solely on domain shifts while maintaining a fixed class set.

However, both settings oversimplify real-world dynamics by restricting increments to a single, predefined type, which limits their practical applicability. In reality, new classes and domains often emerge simultaneously and unpredictably. For instance, the discovery of novel object classes frequently coincides with domain shifts induced by varying weather or lighting conditions in autonomous driving scenarios~\cite{luosheng, abhilash2025enhancing}. To better model such complexities, Versatile Incremental Learning (VIL)~\cite{vil} was recently proposed as a class- and domain-agnostic setting where the model lacks prior knowledge of which classes or domains will appear next, as shown in Fig.~\ref{fig:scenario}. Nevertheless, VIL still assumes a static increment scale (i.e., a fixed number of new classes or domains per task), which deviates significantly from dynamic real-world conditions where task sizes fluctuate unpredictably. To bridge this gap, it is essential to investigate a more general IL paradigm that simultaneously accounts for the randomness of both increment types across tasks and increment scales within each task, thereby offering a more faithful reflection of dynamic increment characteristics within realistic scenarios.

In this work, we introduce \textbf{Universal Incremental Learning (UIL)}, a more realistic IL setting that simultaneously incorporates randomness in both increment type and scale, as conceptually illustrated in Fig.~\ref{fig:scenario}. Under UIL, each incoming task may introduce arbitrary combinations of new classes and/or new domains, while the number of newly introduced classes or domains varies unpredictably across tasks. The term ``universal'' signifies that the model operates without any prior knowledge regarding the type or magnitude of upcoming increments. 
``Universal'' focuses on open-world incremental learning with explicitly defined class labels, but does not involve unknown category discovery in this work.

As formally compared in Table~\ref{tab:comparison_settings}, while all incremental settings inherently face the foundational challenge of Catastrophic Forgetting (CF), they differ substantially in the randomness of increment types and scales. In particular, VIL introduces randomness in increment type but still assumes a fixed increment scale, whereas UIL introduces randomness in both type and scale. Consequently, UIL is fundamentally distinct in its attendant challenges arising from distribution randomness.

Unlike GCL, which operates on sample-level streaming data without explicit task boundaries (CF + Blurry Boundaries), UIL retains explicit task boundaries and batch-level increments. Compared to LT-CIL, which addresses static class-frequency imbalances under a purely class-incremental paradigm (CF + Class Imbalance), UIL introduces a dual challenge: it suffers from the long-tail effect caused by random scales (intra-task confusion), but more critically, it must handle the cross-task feature conflicts caused by the unpredictable mixture of class and domain increments (inter-task confusion).

\begin{table}[!htbp]
	\centering
	\caption{Comparison of UIL with representative Incremental Learning settings.}
	\label{tab:comparison_settings}
	\renewcommand{\arraystretch}{1.3}
	\resizebox{\columnwidth}{!}{
		\begin{tabular}{llcl}
			\toprule
			\textbf{Setting} & \textbf{Increment Characteristics (Type \& Scale)} & \textbf{Task Boundary} & \textbf{Core Challenge} \\
			\midrule
			Traditional IL (CIL/DIL) & Fixed type, Fixed scale & Explicit & Catastrophic Forgetting (CF) \\
			VIL & Random type, Fixed scale & Explicit & CF + Type Randomness \\
			GCL / Task-free CL & Random type, Continuous stream & No / Blurry & CF + Blurry Boundaries \\
			LT-CIL & Class only, Imbalanced scale & Explicit & CF + Class Imbalance \\
			\textbf{UIL (Ours)} & \textbf{Random type, Random scale} & \textbf{Explicit} & \textbf{CF + Inter/Intra-task Confusion} \\
			\bottomrule
		\end{tabular}
	}
\end{table}

This setting introduces two fundamental challenges that existing IL methods struggle to address:
\begin{enumerate}[label=\arabic*]
	\item \textit{Inter-task distribution confusion:} Since increment types vary randomly across sequential tasks, designing specialized modules tailored to a specific type (e.g., purely class- or domain-incremental) becomes ineffective~\cite{l2p,sprompts}. Furthermore, when the model encounters a novel increment type, it tends to learn knowledge that conflicts with previously acquired representations. This conflict hinders accurate and deterministic predictions, ultimately exacerbating catastrophic forgetting.
    \item \textit{Intra-task distribution confusion:} Due to the random scale of increments, the cumulative class distribution inevitably becomes long-tailed, as different classes emerge at varying frequencies in realistic scenarios.
    Standard optimizers~\cite{sgd,adam} compute batch-averaged gradients, which inherently bias updates toward head classes with abundant samples. Consequently, gradient updates become asymmetrical: head classes dominate the optimization trajectory while tail classes receive insufficient gradient signals, leading to underfitting on rare classes and overfitting on frequent ones.

\end{enumerate}

To tackle these dual challenges, we propose \textbf{MiCo} (\textbf{Mi}tigating \textbf{Co}nfusion), a streamlined yet highly effective framework tailored for UIL. MiCo integrates a multi-objective learning scheme with two decoupled recalibration modules that operate on gradient direction and magnitude, respectively. Specifically, to combat inter-task confusion, we jointly optimize a cross-entropy loss $\mathcal{L}_{\mathrm{ce}}$ and an entropy minimization loss $\mathcal{L}_{\mathrm{em}}$~\cite{tent,niu2023towards}. While $\mathcal{L}_{\mathrm{ce}}$ aligns predictions with ground-truth labels, $\mathcal{L}_{\mathrm{em}}$ explicitly reduces the Prediction Distribution Entropy (PDE) of each sample, forcing the model to produce sharp, deterministic predictions regardless of shifting increment types. To further stabilize this multi-objective optimization, we introduce a direction recalibration module that detects and resolves conflicting gradient updates (i.e., Gradient Cosine Similarity, $\mathrm{GCS} < 0$), ensuring smoother convergence across tasks. To address intra-task confusion, we design a magnitude recalibration module that rescales gradient magnitudes based on the observed positive correlation between class frequency and gradient strength. This principled rescaling mitigates asymmetric optimization, ensuring balanced learning across head and tail classes. The main contributions of this work are summarized as follows:
\begin{itemize}
	\item We formally investigate Universal Incremental Learning (UIL), a more general and realistic IL setting that simultaneously accounts for randomness in both the type and scale of incoming increments.
	\item We propose MiCo, a novel framework that effectively mitigates inter- and intra-task distribution confusion in UIL via a multi-objective learning scheme coupled with direction- and magnitude-decoupled gradient recalibration.
	\item Extensive experiments across four benchmarks demonstrate that MiCo significantly outperforms state-of-the-art methods under both the proposed UIL and the existing VIL scenarios.
\end{itemize}

\section{Related Work}
\label{sec:related_work}

\begin{figure}[]
	\centering
	\includegraphics[width=0.95\columnwidth]{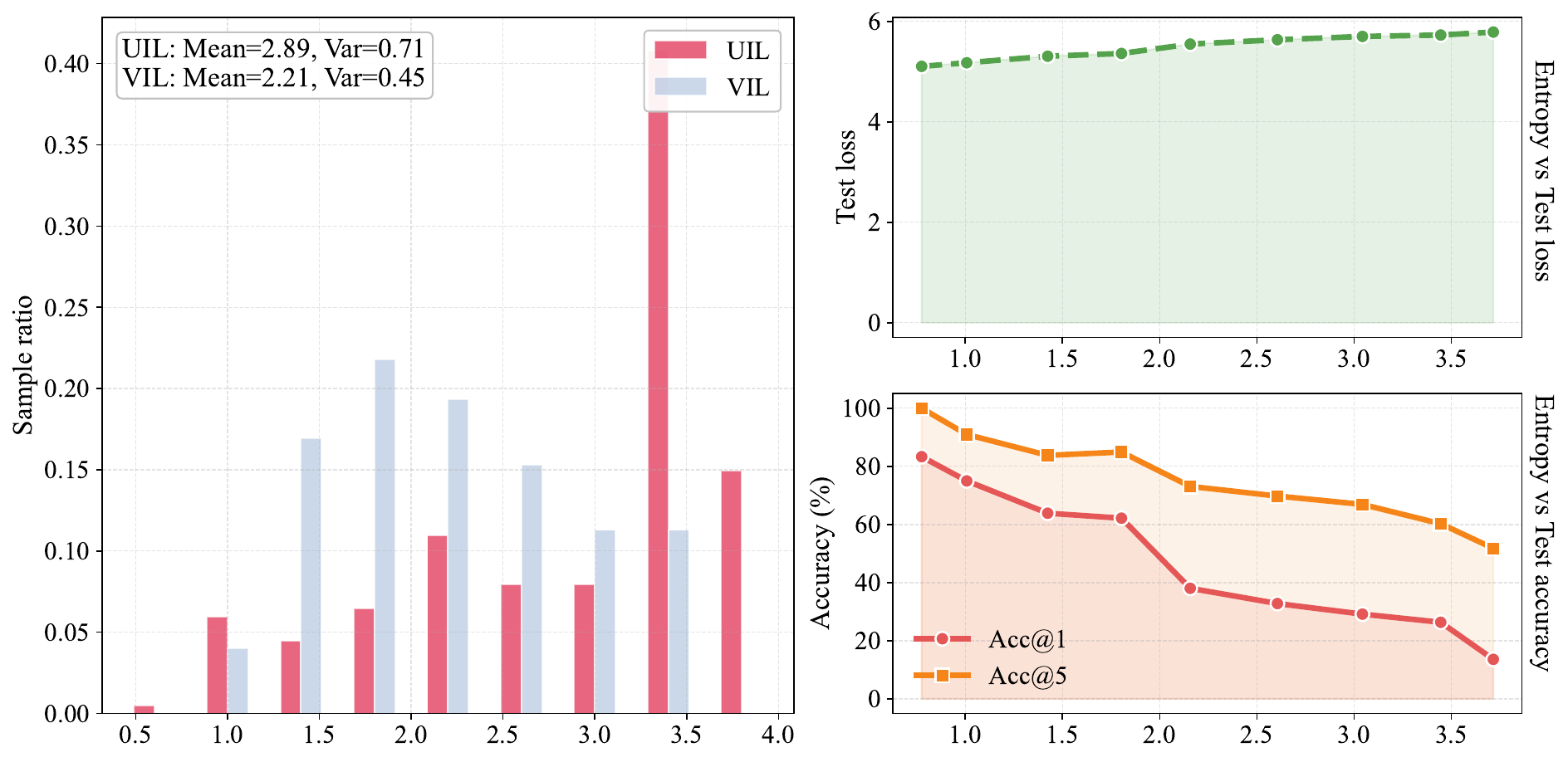}
	\caption{Analysis of prediction distribution entropy (PDE) in the proposed UIL scenario and its correlation with test loss/accuracy. \textbf{Left:} Comparison of PDE distributions between UIL and VIL. UIL exhibits higher mean and variance (UIL: Mean=2.89, Var=0.71; VIL: Mean=2.21, Var=0.45), indicating greater prediction confusion. \textbf{Right:} Samples grouped by entropy intervals during inference. Lower-entropy samples consistently yield lower test loss and higher accuracy, validating entropy minimization as an effective strategy for improving model reliability.}
	\label{fig:entropy}
\end{figure}

\subsection{Incremental Learning}
Incremental Learning (IL) fundamentally aims to continuously acquire new knowledge while mitigating catastrophic forgetting, which refers to the tendency of neural networks to overwrite previously learned representations when exposed to new data.
To address this challenge, mainstream approaches fall into three primary categories. First, \textit{rehearsal-based methods} preserve historical knowledge by replaying stored exemplars~\cite{replay1,dgr} or generating synthetic samples~\cite{gen_cll} to approximate past data distributions. Second, \textit{regularization-based methods}~\cite{vil,zscl,du2026negative} enforce parameter stability by penalizing updates to weights deemed critical for prior tasks, thereby balancing plasticity and stability. Third, with the advent of large-scale pre-trained models~\cite{learn_to_prompt}, \textit{prompt-based methods} have emerged as a highly efficient alternative, leveraging dynamic prompt selection~\cite{l2p}, complementary prompt tuning~\cite{dualprompt}, and attention-based decomposition~\cite{codaprompt} to adapt frozen backbones without modifying core representations. While effective under controlled settings (e.g., fixed class or domain increments), these paradigms typically rely on predefined task structures or static increment scales, limiting their adaptability to the unpredictable dynamics inherent in UIL. 

Recent advances have further explored buffer-free continual strategies~\cite{gupta2025buffer, misa}, task-free adaptation via expansible architectures~\cite{ye2026online}, and feature-space expansion mechanisms tailored for long-tailed class distributions~\cite{wu2025feature}. In particular, MISA~\cite{misa} introduces a replay-independent prompt-based framework for General Continual Learning (GCL) that mitigates catastrophic forgetting through a non-parametric logit mask. However, it is crucial to distinguish UIL from GCL and task-free CL: while GCL typically operates on sample-level streaming data without explicit task boundaries, UIL \textbf{retains explicit task boundaries and batch-level increments}. The ``Universal'' in UIL signifies a zero-prior-knowledge assumption regarding the \textit{type} and \textit{scale} of increments within each defined task boundary, rather than a task-free setting.

Another closely related line of work is Long-Tailed Class-Incremental Learning (LT-CIL), which addresses static class-frequency imbalances. Recent advances such as PSR~\cite{psr}, GPA~\cite{gpa}, and Sub-prototype~\cite{subprototype} propose proactive soft-orthogonal regulation, geometric prototype alignment, and independent sub-prototype construction, respectively. While effective in pure CIL, these methods inherently rely on fixed task scales and known class boundaries during the base phase. Consequently, their proactive mechanisms and geometric alignment assumptions become inapplicable under UIL's random class/domain mixing and unpredictable increment scales.

Furthermore, to handle single-type randomness, recent state-of-the-art methods like CL-LoRA~\cite{cl_lora} and DUCT~\cite{duct} have been proposed for rehearsal-free class-incremental and domain-incremental learning, respectively. Although they represent significant progress in handling class or domain shifts individually, they lack the explicit mechanisms to address the \textit{dual} distribution randomness (both type and scale) inherent in UIL, underscoring the necessity for a more universal framework tailored to our setting.

\begin{figure}[]
	\centering
	\includegraphics[width=0.9\columnwidth]{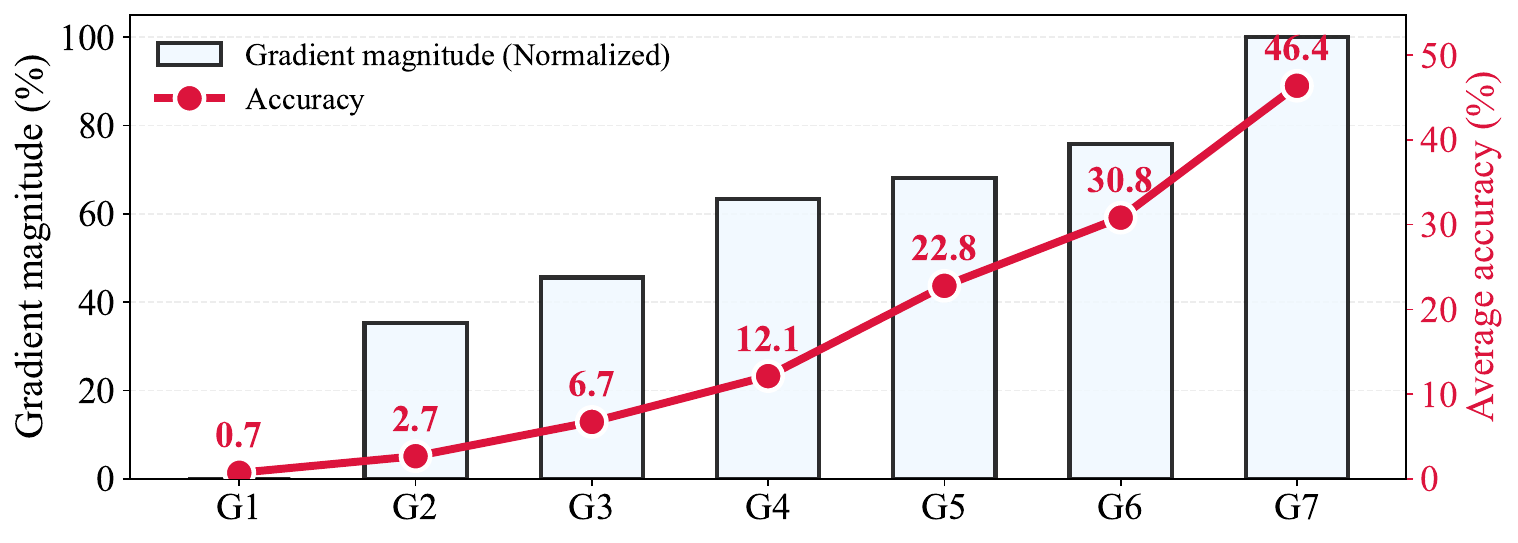}
	\caption{Correlation between normalized gradient magnitude and average accuracy across class groups G1--G7 (grouped by emergence order). Blue bars denote normalized gradient magnitude; the red dashed line denotes average accuracy (\%). The clear positive trend indicates that classes receiving larger gradient updates achieve higher accuracy, revealing an optimization bias toward head classes under imbalanced distributions.}
	\label{fig:spearman}
\end{figure}

\subsection{Entropy Minimization}
Entropy minimization has long been established as a powerful regularizer for improving model calibration and generalization under distribution shifts. In source-free domain adaptation~\cite{guo2021unsupervised}, as well as test-time adaptation~\cite{tent} and adaptive model merging~\cite{chen2025toward,qiang2025split}, minimizing the entropy of predictive distributions encourages models to produce confident outputs on shifted or unlabeled data. In the context of UIL, the model faces severe prediction uncertainty due to the random emergence of novel classes and domains across sequential tasks. By explicitly minimizing the Prediction Distribution Entropy (PDE), our approach enforces sharp, deterministic predictions regardless of fluctuations in increment types, thereby directly mitigating the \textit{inter-task distribution confusion} analyzed in this work.

\subsection{Optimization for Incremental Learning}
Gradient manipulation has proven effective in navigating the stability-plasticity trade-off during continual learning. Optimization-based IL methods primarily focus on controlling gradient directions to prevent interference, with representative works employing orthogonal subspace projection~\cite{ogd} and feasible-space gradient constraints~\cite{gpm}. Specifically, GEM~\cite{gem} and A-GEM~\cite{agem} project current task gradients onto feasible spaces defined by replayed samples, ensuring updates do not degrade prior performance. DGR~\cite{dgr} additionally incorporates gradient magnitude rectification alongside knowledge distillation from stored replay data. However, these methods predominantly assume relatively balanced or stationary task distributions. In dynamic settings like UIL, rigidly aligning gradients with historical directions often induces severe optimization conflicts when increment types vary unpredictably~\cite{vil}. Moreover, they lack explicit mechanisms to address the \textit{intra-task} gradient magnitude asymmetry caused by randomly evolving, long-tailed class distributions. Complementary to these gradient-based approaches, recent optimization-focused studies have introduced low-redundancy distillation techniques~\cite{liu2025low} and dynamic masking-guided frameworks for online continual learning~\cite{liang2025masking}, which explicitly address representation conflicts under streaming data. However, they largely treat gradient direction and magnitude as coupled updates, lacking explicit decoupled recalibration mechanisms for the dual distribution randomness inherent in UIL. Our proposed direction and magnitude recalibration modules are specifically designed to resolve these dual optimization bottlenecks without relying on memory buffers or restrictive projection constraints.

\begin{figure*}[]
	\centering
	\includegraphics[width=0.9\linewidth]{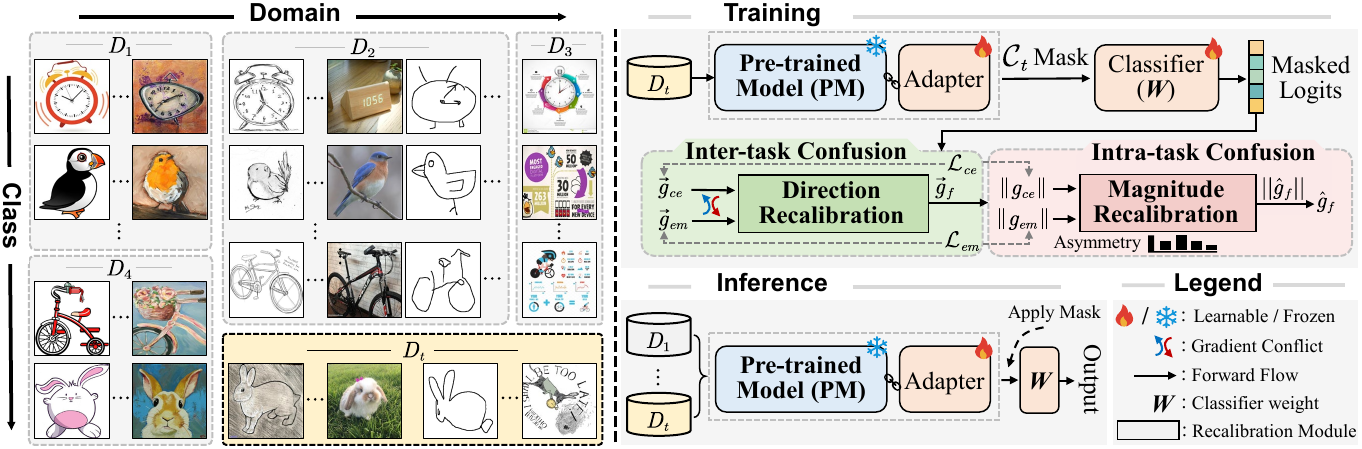}
	\caption{Overview of the proposed UIL scenario (left) and the MiCo framework (right). During training on the current task $\mathcal{D}_{t}$, MiCo employs a multi-objective learning scheme ($\mathcal{L}_{ce} + \gamma\mathcal{L}_{em}$) to enforce accurate and deterministic predictions. A \textit{direction recalibration} module mitigates gradient conflicts between $\vec{g}_{ce}$ and $\vec{g}_{em}$, while a \textit{magnitude recalibration} module rescales gradient norms (e.g., $\lVert \boldsymbol{g}_{ce} \rVert$, $\lVert \boldsymbol{g}_{em} \rVert$) to alleviate asymmetrical optimization induced by imbalanced class distributions. The recalibrated gradient $\hat{\boldsymbol{g}}_{f}$, combining updated direction $\vec{g}_{f}$ and magnitude $\lVert \hat{\boldsymbol{g}}_{f} \rVert$, is used for parameter updates. During inference, the model is evaluated on all accumulated tasks $\mathcal{D}_{\text{test}} = \{\mathcal{D}_{1}, \dots, \mathcal{D}_{t}\}$.
		Note: A cumulative known-class binary mask is applied to logits. During training, classes outside the current task are set to $-\infty$; during inference, the mask contains all classes learned so far and does not require oracle task identity.}
	\label{fig:overview}
\end{figure*}

\section{Universal Incremental Learning}
\label{sec:background}

\subsection{Problem Formulation and Definition}
\label{subsec:problem_setup}

\noindent\textbf{Problem Formulation.}
We formally define Universal Incremental Learning (UIL). Let $\mathcal{D} = \{D_{1}, D_{2}, \dots, D_{T}\}$ denote a sequence of $T$ disjoint tasks. Each task $D_{t} = \{\mathcal{X}_{t}, \mathcal{Y}_{t}\}$ comprises $n_{t}$ images $\mathcal{X}_{t} = \{x_{1}, x_{2}, \dots, x_{n_{t}}\}$ and their corresponding labels $\mathcal{Y}_{t} = \{y_{1}, y_{2}, \dots, y_{n_{t}}\}$. For $i \in [1, 2, \dots, n_{t}]$, each label $y_{i} = (c_{i}, m_{i})$ is annotated with a class index $c_{i} \in \mathcal{C}_{t}$ and a domain index $m_{i} \in \mathcal{M}_{t}$. Formally, for any $t, t' \in [1, 2, \dots, T]$ with $t \neq t'$, disjoint tasks satisfy $\mathcal{X}_{t} \cap \mathcal{X}_{t'} = \emptyset$ and $\mathcal{Y}_{t} \cap \mathcal{Y}_{t'} = \emptyset$.

UIL simultaneously accounts for randomness in both the \textit{type} and \textit{scale} of increments. Consequently, the cardinalities of the class set $\mathcal{C}_{t}$ and domain set $\mathcal{M}_{t}$, denoted as $\lVert \mathcal{C}_{t} \rVert$ and $\lVert \mathcal{M}_{t} \rVert$, are random variables. The model $f^{\theta}_{T}$, parameterized by $\theta_{T}$ at task $T$, is trained solely on the current task's training set $D_{\text{train}} = D_{T}$, yet evaluated on the cumulative test set $D_{\text{test}} = \{D_{1}, D_{2}, \dots, D_{T}\}$ to assess overall performance across all $T$ tasks. This objective is formalized as:
\begin{equation}
	\label{eq:obj}
	\mathcal{L}_{\theta} = \sum_{t = 1}^{T} \mathbb{E}_{(x, y) \sim (\mathcal{X}_{t}, \mathcal{Y}_{t})} \left[ \ell\big(f^{\theta}_{t}(x), y\big) \right],
\end{equation}
where $\ell(\cdot, \cdot)$ denotes the standard cross-entropy loss for classification.

Notably, the universality of UIL is evident when considering $\|\mathcal{C}_t\|$ and $\|\mathcal{M}_t\|$ at task $t$.
The distinction between UIL and existing settings can be formally characterized through the increment state $\mathcal{S}_t = (\tau_t, k_t)$ at task $t$, where $\tau_t \in \{\mathrm{class}, \mathrm{domain}, \mathrm{mixed}\}$ represents the increment type and $k_t \in \mathbb{N}^+$ represents the increment scale. Based on this formulation, existing settings are characterized as follows: (i) CIL/DIL: both $\tau_t$ and $k_t$ are deterministic constants; (ii) VIL: $\tau_t$ is a random variable while $k_t$ remains deterministic; (iii) UIL: both $\tau_t$ and $k_t$ are treated as unknown random variables without any prior structural assumption. The maximum-entropy interpretation simply characterizes this zero-prior-knowledge condition. This formalization shows that VIL is a special case of UIL where $\mathrm{Var}(k_t) = 0$. Since UIL introduces joint randomness in both increment type and scale, it induces strictly higher expected prediction uncertainty than VIL, as the model must cope with this joint randomness without structural priors.

\noindent\textbf{Definition 3.1 (Prediction Distribution Entropy, PDE).}
At task $t$, let $\hat{y}_{i,c}$ denote the logit of the $i$-th sample $x_{i}$ for the $c$-th class, as predicted by $f^{\theta}_{t}$. The corresponding probability after softmax is $p(\hat{y}_{i,c})$. The PDE of $x_{i}$ is defined as:
\begin{equation}
	\label{eq:entropy}
	H(y_i) = -\sum_{c=1}^{\lVert \mathcal{C}_{t} \rVert} p(\hat{y}_{i,c}) \log p(\hat{y}_{i,c}).
\end{equation}
As shown in Eq.~\ref{eq:entropy}, $H(y_i)$ achieves its maximum value $\log \lVert \mathcal{C}_{t} \rVert$ when $p(\hat{y}_{i})$ follows a uniform distribution (indicating maximal confusion among classes in $\mathcal{C}_{t}$), and its minimum value $0$ when $p(\hat{y}_{i})$ follows a one-hot distribution (indicating a fully deterministic prediction):
\begin{equation}
	\label{eq:entropy_value}
	0 \leq H(y_{i}) \leq \log \lVert \mathcal{C}_{t} \rVert.
\end{equation}

\noindent\textbf{Definition 3.2 (Gradient Cosine Similarity, GCS).}
Let $\theta_{u,v}$ denote the angle between two gradient vectors $g_{u}$ and $g_{v}$. The GCS between them is defined as:
\begin{equation}
	\label{eq:gcs}
	\cos\theta_{u,v} = \frac{g_{u}}{\lVert g_{u} \rVert} \cdot \frac{g_{v}}{\lVert g_{v} \rVert},
\end{equation}
where conflicting gradients occur when $\cos\theta_{u,v} < 0$.

\subsection{Challenge Analysis}
\label{subsec:scenario_challenge}

To systematically analyze the challenges inherent to UIL, we conduct a two-level empirical investigation.

\noindent\textbf{Inter-task distribution confusion.}
At the inter-task level, we compute the PDE of all test samples under both UIL and VIL scenarios after incremental training on $T=30$ sequential tasks using DomainNet~\cite{DomainNet}, which contains diverse classes and domains. We partition test samples into ascending entropy intervals $\mathcal{I} = \{\mathcal{I}_{1}, \mathcal{I}_{2}, \dots, \mathcal{I}_{27}\}$ based on $H(y_i)$ and compute the sample ratio per interval. As illustrated in Fig.~\ref{fig:entropy}, UIL induces significantly greater prediction confusion than VIL, evidenced by: (1) a markedly smaller proportion of samples in low-entropy intervals, and (2) a relatively larger proportion in high-entropy intervals. Furthermore, the right panel of Fig.~\ref{fig:entropy} demonstrates a strong negative correlation between entropy and test loss, and a positive correlation with accuracy. This empirically validates that minimizing prediction entropy is an effective strategy for mitigating \textit{inter-task distribution confusion} caused by random increment types.

\noindent\textbf{Intra-task distribution confusion.}
At the intra-task level, we analyze the impact of random increment scales on optimization dynamics. During training, we compute the average gradient magnitude of each class's weight vector across three random seeds; during inference, we compute the corresponding per-class test accuracy. The joint relationship among class index, gradient magnitude, and accuracy is visualized in Fig.~\ref{fig:spearman}. We observe a pronounced positive correlation: classes with larger gradient magnitudes consistently achieve higher accuracy.
In the absence of explicit intervention, this leads to \textit{asymmetric optimization}: gradient updates disproportionately favor head classes with abundant samples, while tail classes receive insufficient optimization signals.
Consequently, the model becomes biased toward predicting head classes and underfits tail classes, failing to achieve balanced performance across the long-tailed distribution induced by random increment scales. We term this phenomenon \textit{intra-task distribution confusion}.

\section{Methodology}
\label{sec:method}

Guided by the empirical analysis in Sec.~\ref{sec:background}, we propose \textbf{MiCo} (\textbf{Mi}tigating \textbf{Co}nfusion), a streamlined framework tailored for UIL. As illustrated in Fig.~\ref{fig:overview}, MiCo maintains a frozen pre-trained backbone and a trainable classifier head during training on the current task $\mathcal{D}_{t}$, while keeping all parameters frozen for evaluating the cumulative test set $\mathcal{D}_{\text{test}} = \{\mathcal{D}_{1}, \dots, \mathcal{D}_{t}\}$. To address \textit{inter-task distribution confusion}, Sec.~\ref{subsec:inter-task} introduces a multi-objective learning scheme that jointly optimizes $\mathcal{L}_{ce}$ and $\mathcal{L}_{em}$ to enforce sharp, deterministic predictions. We further augment this with a \textit{direction recalibration} module that explicitly resolves conflicting gradient directions (e.g., $\vec{g}_{ce} = \boldsymbol{g}_{ce}/\lVert \boldsymbol{g}_{ce} \rVert$ and $\vec{g}_{em} = \boldsymbol{g}_{em}/\lVert \boldsymbol{g}_{em} \rVert$). To mitigate \textit{intra-task distribution confusion}, Sec.~\ref{subsec:intra-task} presents a \textit{magnitude recalibration} module that dynamically rescales gradient norms to counteract asymmetrical optimization. Crucially, this decoupled recalibration strategy implicitly balances the stability-plasticity trade-off, offering a principled optimization pathway to mitigate catastrophic forgetting while adapting to realistic increments. The final recalibrated gradient $\hat{\boldsymbol{g}}_{f}$ is then used for backward propagation.

\subsection{Mitigating Confusion from Inter-task Distribution Randomness}
\label{subsec:inter-task}

In conventional CIL or DIL settings, incremental tasks follow a predefined structure (e.g., exclusively new classes or domains), allowing the model to learn type-specific representations with prior knowledge. In contrast, UIL operates without any prior knowledge of increment types, causing newly introduced knowledge to frequently conflict with previously learned representations. This conflict manifests as degraded prediction confidence and accelerates catastrophic forgetting. To mitigate this \textit{inter-task distribution confusion} and alleviate knowledge degradation, we design a multi-objective learning scheme that jointly optimizes accuracy and prediction certainty.

Unlike conventional IL with known increment types, UIL's ``type-unknown'' condition causes severe decision boundary blurring (Fig.~\ref{fig:entropy}). Here, $\mathcal{L}_{em}$ acts as a Type-agnostic anchor, forcing sharp predictions against random shifts. Crucially, to prevent over-confidence, $\mathcal{L}_{em}$ is synergistically coupled with the Direction Recalibration (DIR) module. DIR functions as an optimization constraint: if $\mathcal{L}_{em}$ pushes the model toward an incorrect, overconfident minimum, DIR detects the resulting gradient conflict ($\mathrm{GCS} < 0$) and projects the update away from the conflicting subspace, neutralizing the over-confidence risk.

As illustrated in Fig.~\ref{fig:overview}, we jointly train the model using the standard cross-entropy loss $\mathcal{L}_{ce}$ and an entropy minimization loss $\mathcal{L}_{em}$, a technique widely adopted in test-time adaptation~\cite{tent,niu2023towards}. For an input sample $x_i$, let $\hat{y}_{i,c}$ denote the logit for the $c$-th class, and $p(\hat{y}_{i,c})$ be the corresponding softmax probability. The prediction distribution entropy (PDE) of $x_i$, denoted $H(y_i)$, quantifies prediction uncertainty (Eq.~\ref{eq:entropy}). By simultaneously minimizing the classification error and the PDE across tasks with random increment types, we enforce the model to produce sharp, deterministic predictions regardless of the underlying distribution shift. The overall multi-objective loss is formulated as:
\begin{equation}
	\label{eq:tot_loss}
	\mathcal{L} = \mathcal{L}_{ce} + \gamma \mathcal{L}_{em},
\end{equation}
where $\gamma$ balances the two objectives (see Sec.~\ref{subsec:ablation_study} for sensitivity analysis). This linear combination intrinsically aligns with UIL by countering unpredictable distribution shifts: $\mathcal{L}_{em}$ acts as a curvature-aware regularizer that smooths the loss landscape toward flatter minima, thereby fostering robust, task-agnostic generalization across arbitrary increments without relying on structural priors.

While multi-objective learning improves prediction certainty, optimizing $\mathcal{L}_{ce}$ and $\mathcal{L}_{em}$ simultaneously can introduce conflicting gradient directions (i.e., $\cos \theta_{ce, em} < 0$). To resolve this, we introduce a \textit{direction recalibration} module. Let $\boldsymbol{g}^{c}_{ce}$ and $\boldsymbol{g}^{c}_{em}$ be the gradients of the $c$-th classifier weight vector w.r.t. $\mathcal{L}_{ce}$ and $\mathcal{L}_{em}$, respectively. Their unit directions are $\vec{g}^{c}_{ce} = \boldsymbol{g}^{c}_{ce}/\lVert \boldsymbol{g}^{c}_{ce} \rVert$ and $\vec{g}^{c}_{em} = \boldsymbol{g}^{c}_{em}/\lVert \boldsymbol{g}^{c}_{em} \rVert$. The initial combined direction is $\vec{g}^{c} = \vec{g}^{c}_{ce} + \gamma \vec{g}^{c}_{em}$. We then apply a conflict-aware adjustment:
\begin{equation}
	\label{eq:direction}
	\begin{aligned}
		\vec{g}^{c} &= \vec{g}_{ce}^{c} + \gamma \vec{g}_{em}^{c}, \\
		\vec{g}_{f}^{c} &= \vec{g}^{c} + \beta \vec{g}_{w}^{c},
	\end{aligned}
\end{equation}
where $\beta$ is a hyper-parameter controlling the recalibration strength, and $\vec{g}_{w}^{c}$ is the conflict-mitigation offset derived from CAGrad~\cite{cadgrad}. Specifically, CAGrad formulates gradient alignment as the following constrained optimization problem:
\begin{equation}
	\label{eq:cagrad_opt}
	\min_{\boldsymbol{d}} \max_{k \in \{ce, em\}} \langle \boldsymbol{g}^{c}_{k}, \boldsymbol{d} \rangle \quad \text{s.t.} \quad \lVert \boldsymbol{d} - \bar{\boldsymbol{g}}^{c} \rVert_2 \le \lambda \lVert \bar{\boldsymbol{g}}^{c} \rVert_2, \quad \lambda \in [0,1],
\end{equation}
where $\bar{\boldsymbol{g}}^{c} = (\boldsymbol{g}^{c}_{ce} + \boldsymbol{g}^{c}_{em})/2$ and $\lambda \in [0,1]$ bounds the deviation from the average descent direction. Solving the dual problem via KKT conditions yields a closed-form update that projects the gradient away from the conflicting subspace ($\cos\theta_{ce,em}<0$). In our implementation, the difference between this CAGrad-optimal direction and the naive average $\vec{g}^{c}$ is encapsulated as the offset $\vec{g}_{w}^{c}$, which guarantees non-negative inner products with both $\boldsymbol{g}^{c}_{ce}$ and $\boldsymbol{g}^{c}_{em}$ whenever the trust region permits. This conflict-aware alignment helps stabilize the optimization trajectory, reducing the risk of aggressively overwriting directions critical for prior tasks. The hyper-parameter $\beta$ in Eq.~\ref{eq:direction} linearly scales $\vec{g}_{w}^{c}$ to explicitly trade off conflict resolution against optimization step size.

\begin{figure}[]
	\centering
	\includegraphics[width=1.0\columnwidth]{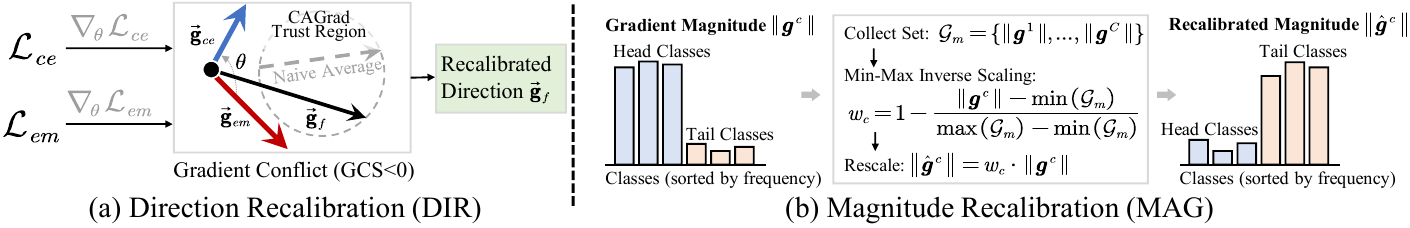}
	\caption{Internal mechanisms of the MiCo modules. (a) DIR resolves gradient conflicts between $\mathcal{L}_{ce}$ and $\mathcal{L}_{em}$ via CAGrad-based projection to obtain the recalibrated direction $\vec{g}_f$. (b) MAG rebalances asymmetric class-wise gradient magnitudes through min-max inverse scaling to amplify tail-class signals.}
	\label{fig:dir_mag_mechanism}
\end{figure}

\subsection{Mitigating Confusion from Intra-task Distribution Randomness}
\label{subsec:intra-task}

In realistic settings, the random scale of increments leads to highly imbalanced class distributions, as new classes emerge at varying frequencies. Standard optimizers~\cite{sgd,adam} compute batch-averaged gradients, which inherently bias updates toward head classes with abundant samples. This results in \textit{asymmetrical optimization}: weight vectors for head classes receive disproportionately large gradient magnitudes, while tail classes suffer from insufficient optimization signals, leading to over-fitting on frequent classes and under-fitting on rare ones. In continual learning, such optimization bias can disproportionately harm the retention of underrepresented classes, thereby accelerating forgetting. To counteract this \textit{intra-task distribution confusion}, we propose a \textit{magnitude recalibration} module that dynamically rescales gradient norms based on class frequency. Let $\mathcal{G}_{m} = \{\lVert \boldsymbol{g}^{1} \rVert, \lVert \boldsymbol{g}^{2} \rVert, \dots, \lVert \boldsymbol{g}^{C} \rVert\}$ denote the set of gradient magnitudes for all $C$ classifier weight vectors. The magnitude for the $c$-th class is computed from the combined gradient vector:
\begin{equation}
	\label{eq:g_i}
	\lVert \boldsymbol{g}^{c} \rVert = \lVert \boldsymbol{g}^{c}_{ce} + \gamma \boldsymbol{g}^{c}_{em} \rVert,
\end{equation}
where $\gamma$ matches the coefficient in Eq.~\ref{eq:tot_loss}. We then apply a class-specific scaling factor $w_c$ derived from min-max normalization over $\mathcal{G}_{m}$:
\begin{equation}
	\label{eq:magnitude}
	\begin{aligned}
		\lVert \hat{\boldsymbol{g}}^{c} \rVert &= w_{c} \cdot \lVert \boldsymbol{g}^{c} \rVert, \\
		\text{where} \quad w_{c} &= 1 - \frac{\lVert \boldsymbol{g}^{c} \rVert - \min(\mathcal{G}_{m})}{\max(\mathcal{G}_{m}) - \min(\mathcal{G}_{m}) + \epsilon}.
	\end{aligned}
\end{equation}

\begin{algorithm}[t]
	\caption{Training Procedure of MiCo Framework}
	\label{alg:mico}
	\begin{algorithmic}[1]
		\Require Task data $\mathcal{D}_t$, classifier weights $\boldsymbol{W}$, hyperparameters $\gamma, \beta, \lambda, \epsilon$
		\Ensure Updated $\boldsymbol{W}$
		\For{each mini-batch $(\boldsymbol{X}, \boldsymbol{Y}) \in \mathcal{D}_t$}
		\State Compute losses: $\mathcal{L}_{ce}$ and $\mathcal{L}_{em} = -\frac{1}{|\boldsymbol{X}|}\sum_{i,c} p(c|\boldsymbol{x}_i) \log p(c|\boldsymbol{x}_i)$
		\For{each class $c \in \mathcal{C}_t$}
		\State Compute unit gradients $\vec{g}^{c}_{ce}, \vec{g}^{c}_{em}$ and combined direction $\vec{g}^{c} = \vec{g}^{c}_{ce} + \gamma \vec{g}^{c}_{em}$
		\If{$\langle \vec{g}^{c}_{ce}, \vec{g}^{c}_{em} \rangle < 0$} \Comment{DIR: Resolve conflict}
		\State $\vec{g}^{c}_{f} \leftarrow \vec{g}^{c} + \beta \cdot \text{CAGrad}(\vec{g}^{c}_{ce}, \vec{g}^{c}_{em}, \lambda)$
		\Else
		\State $\vec{g}^{c}_{f} \leftarrow \vec{g}^{c}$
		\EndIf
		\State Compute magnitude $\|\boldsymbol{g}^{c}\| = \|\boldsymbol{g}^{c}_{ce} + \gamma \boldsymbol{g}^{c}_{em}\|$ and collect $\mathcal{G}_m$
		\State MAG: Rescale magnitude $\|\hat{\boldsymbol{g}}^{c}\| \leftarrow \left(1 - \frac{\|\boldsymbol{g}^{c}\| - \min(\mathcal{G}_m)}{\max(\mathcal{G}_m) - \min(\mathcal{G}_m) + \epsilon}\right) \|\boldsymbol{g}^{c}\|$
		\State Reconstruct gradient: $\hat{\boldsymbol{g}}^{c}_{f} \leftarrow \|\hat{\boldsymbol{g}}^{c}\| \cdot \vec{g}^{c}_{f}$
		\EndFor
		\State Update weights: $\boldsymbol{W} \leftarrow \boldsymbol{W} - \eta \cdot [\hat{\boldsymbol{g}}^{1}_{f}; \dots; \hat{\boldsymbol{g}}^{C}_{f}]$
		\EndFor
		\State \Return Updated $\boldsymbol{W}$
	\end{algorithmic}
\end{algorithm}

Here, $\epsilon$ (e.g., $10^{-8}$) is a small stability constant introduced to strictly prevent division by zero when gradient magnitudes are identical ($\max = \min$), ensuring absolute numerical stability. This formulation inversely scales gradient magnitudes: classes with larger accumulated gradients (typically head classes) receive smaller weights $w_c \in [0, 1]$, while tail classes receive larger weights, effectively balancing the optimization landscape. The step-by-step optimization process of MiCo is detailed in Algorithm~\ref{alg:mico}.

In UIL's long-tailed distributions, large gradients are overwhelmingly driven by class frequency rather than intrinsic hardness (Fig.~\ref{fig:spearman}), representing redundant optimization dominance from head classes. Conversely, tail classes inherently possess small gradients due to data scarcity. MAG's inverse scaling explicitly assigns larger weights to these smaller gradients, effectively amplifying optimization signals for hard/tail samples. Thus, MAG rectifies sample-imbalance bias without penalizing hard samples.

By preventing head-class gradients from dominating the update step, this mechanism ensures that tail and historical classes receive sufficient gradient signals to maintain their decision boundaries. Finally, the fully recalibrated gradient for the $c$-th weight is reconstructed as $\hat{\boldsymbol{g}}^{c}_{f} = \lVert \hat{\boldsymbol{g}}^{c} \rVert \cdot \vec{g}^{c}_{f}$, combining the calibrated direction $\vec{g}^{c}_{f}$ (Eq.~\ref{eq:direction}) and magnitude $\lVert \hat{\boldsymbol{g}}^{c} \rVert$ (Eq.~\ref{eq:magnitude}). This gradient is then used for backward propagation, enabling MiCo to fairly acquire knowledge across all classes without being dominated by head-class gradients.

\subsection{Theoretical Intuition on Forgetting Mitigation}
\label{subsec:theoretical}
MiCo fundamentally mitigates catastrophic forgetting through principled optimization mechanisms. 

First, the DIR module provides a local optimization stability analysis through surrogate-based gradient alignment. In MiCo, the pre-trained backbone is frozen while the adapter and classifier head remain trainable. Since the adapter is shared across tasks, its updates alter feature representations for both current and historical samples, constituting the primary source of forgetting. The entropy minimization loss $\mathcal{L}_{em}$ enforces deterministic predictions on the current task, which implicitly encourages representation stability shared across tasks under this shared-adapter architecture, making $\boldsymbol{g}_{em}$ a task-agnostic surrogate for representation stability, rather than a direct historical-task gradient. DIR ensures $\langle \vec{g}_f, \vec{g}_{em} \rangle \geq 0$, aligning the update direction with this surrogate constraint. Under a first-order Taylor expansion, this constrains the update direction associated with representation drift and reduces the risk of harmful changes in historical representations. Additionally, the CAGrad trust region constraint $\|\boldsymbol{d} - \bar{\boldsymbol{g}}\| \leq \lambda\|\bar{\boldsymbol{g}}\|$ limits the deviation of the update from the average descent direction, thereby helping control local representation drift. We note that this analysis should be interpreted as a local optimization stability analysis under first-order approximation, rather than an exact guarantee of zero forgetting. In the buffer-free setting of UIL, explicit historical gradients are inaccessible without replay data; the surrogate-based stability constraint via $\mathcal{L}_{em}$ and the trust region together provide an effective forgetting mitigation mechanism through gradient-level optimization without replay data.

Second, the MAG module functions as a dynamic diagonal preconditioner. Long-tailed distributions induce a high condition number in the gradient covariance matrix, causing optimization bias. By applying inverse scaling weights $w_c \propto 1/\|\boldsymbol{g}^c\|$, MAG compresses the eigenvalue spectrum of the effective Hessian. This balances convergence rates across class-specific subspaces, helping alleviate tail-class underfitting and head-class overfitting, thereby stabilizing the global decision boundary.

\begin{table}[!htbp]
	\centering
	\caption{Task configuration on three benchmarks in the UIL and VIL scenarios. $T$, $\lVert\mathcal{C}_{t}\rVert$, and $\lVert\mathcal{M}_{t}\rVert$ denote the total number of sequential tasks, the number of classes, and the number of domains within each incremental task, respectively. In the UIL scenario, $UNK$ indicates that the number of classes and domains are unknown due to the random scale of increments. The value with asterisk (*) indicates that we ensure each task covers multiple domains to prevent the UIL scenario from degrading into VIL for iDigits (which has fewer classes and domains), resulting in fewer overall tasks.}
	
	\begin{adjustbox}{width=\linewidth}
		\begin{tabular}{lcccccc}
			\specialrule{.1em}{.1em}{.1em} 
			\multirow[b]{1.8}{*}{Dataset} & \multicolumn{3}{c}{VIL} & \multicolumn{3}{c}{UIL}  \\ 
			\cmidrule(lr){2-4} \cmidrule(lr){5-7} 
			& $T$ & $\left\|\mathcal{C}_{t}\right\|$ & $\left\|\mathcal{M}_{t}\right\|$ & $T$ & $\left\|\mathcal{C}_{t}\right\|$ & $\left\|\mathcal{M}_{t}\right\|$ \\
			\specialrule{.05em}{.1em}{.1em} 
			iDigits   \cite{iDigits}     & 20 & 2  & 1 & 5* & $UNK$ & $UNK$ \\
			CORe50    \cite{CORe50}      & 40 & 10 & 1 & 40 & $UNK$ & $UNK$ \\
			DomainNet \cite{DomainNet}   & 30 & 10 & 1 & 30 & $UNK$ & $UNK$ \\
			ImageNet-R \cite{imagenet-r}   & 20 & 10 & 1 & 20 & $UNK$ & $UNK$ \\
			\specialrule{.1em}{.1em}{.1em} 
		\end{tabular}
		
	\end{adjustbox}
	\label{tab:task_configuration}
\end{table}

\section{Experiments}
\label{sec:experiments}

In this section, we conduct extensive experiments to evaluate the effectiveness of our proposed framework. First, we describe the experimental setup, including datasets, evaluation metrics, and comparison methods in Sec.~\ref{subsec:exp_setup}. Next, we present the main results to demonstrate the superiority of MiCo in Sec.~\ref{subsec:main_results}. Finally, we verify the contribution of each component through comprehensive ablation studies in Sec.~\ref{subsec:ablation_study}.

\begin{table*}[!tbp]
	\centering
	\caption{Comparison with state-of-the-art IL methods on the UIL and VIL scenarios. Results (\%) are reported in terms of AVA, WA, and Forgetting ($F$) as mean $\pm$ standard deviation across three random seeds. The AVA across both scenarios is averaged as Mean Acc (last column). Best and second-best results are marked in \textbf{bold} and \underline{underline}, respectively.}
	\resizebox{0.8\textwidth}{!}{
		\begin{tabular}{ccccccccc@{}}
			\specialrule{.1em}{.1em}{.1em} 
			\multirow[b]{1.8}{*}{\textbf{Method}}  & \multicolumn{3}{c}{\textbf{Versatile-IL (VIL)}}  & \multicolumn{3}{c}{\textbf{Universal-IL (UIL)}}  & \multirow[b]{1.8}{*}{\textbf{Mean Acc}($\uparrow$)}\\ 
			& AVA($\uparrow$)  & WA($\uparrow$)  & F($\downarrow$)  & AVA($\uparrow$)  & WA($\uparrow$)  & F($\downarrow$) \\
			\specialrule{.05em}{.1em}{.1em} 
			& & &\multicolumn{2}{c}{\textbf{iDigits}} & & & & \\ 
			\specialrule{.05em}{.1em}{.1em} 
			\text{Fine-tuning}  & 36.3$\pm$0.1  & 32.6$\pm$1.0  & \underline{21.5$\pm$5.5}  & 46.5$\pm$0.5  & 50.0$\pm$2.4  & 18.4$\pm$13.0  & 41.4$\pm$0.3 \\
			\specialrule{.05em}{.1em}{.1em} 
			EWC \cite{ewc}  & 37.4$\pm$2.1  & 34.4$\pm$2.3  & 28.0$\pm$3.5  & 61.4$\pm$2.5  & 56.5$\pm$2.8  & 20.0$\pm$4.2  & 48.5$\pm$2.3 \\
			LwF \cite{lwf}  & 38.2$\pm$2.0  & 35.1$\pm$2.2  & 26.0$\pm$3.2  & 63.9$\pm$2.4  & 58.8$\pm$2.6  & 19.0$\pm$3.8  & 51.4$\pm$2.2 \\
			L2P \cite{l2p}  & 57.2$\pm$3.0  & 52.1$\pm$3.7  & 32.1$\pm$4.1  & 74.3$\pm$1.3  & 74.1$\pm$3.0  & 18.0$\pm$9.1  & 65.8$\pm$2.2 \\
			DualPrompt \cite{dualprompt}  & 61.8$\pm$1.5  & 59.0$\pm$2.9  &24.9$\pm$3.6  & 74.3$\pm$0.1  & 75.4$\pm$0.4  & \underline{15.4$\pm$4.5}  & 68.1$\pm$0.8 \\
			CODA-Prompt \cite{codaprompt}  & 64.6$\pm$1.8  & 61.0$\pm$3.5  & 23.6$\pm$1.7  & 79.0$\pm$11.2  & 78.0$\pm$12.7  & 18.3$\pm$17.0  & 71.8$\pm$6.5 \\
			DGR \cite{dgr}  & 65.3$\pm$3.2  & 63.9$\pm$2.6  & 38.0$\pm$3.7  & 79.3$\pm$4.9  & 77.6$\pm$8.1  & 25.4$\pm$5.3  & 72.3$\pm$4.1 \\
			LAE \cite{lae}  & 50.1$\pm$4.3  & 42.5$\pm$5.4  & 36.1$\pm$1.0  & 74.5$\pm$2.2  & 74.4$\pm$1.6  & 21.7$\pm$2.0  & 62.3$\pm$3.3\\
			ICON \cite{vil}  & \underline{68.0$\pm$2.0}  & \underline{65.1$\pm$2.5}  & \textbf{21.3$\pm$2.7}  & \underline{81.8$\pm$8.0}  & \underline{80.2$\pm$10.3}  & 15.7$\pm$13.1  & \underline{74.9$\pm$5.0} \\
			MISA~\cite{misa}  & 67.5$\pm$2.2  & 64.5$\pm$2.8  & 22.0$\pm$3.1  & 80.9$\pm$8.5  & 79.4$\pm$10.1  & 16.5$\pm$11.2  & 74.2$\pm$5.4 \\
			PSR~\cite{psr} & 67.2$\pm$2.4 & 64.5$\pm$2.6 & 25.6$\pm$3.1 & 80.5$\pm$7.5 & 79.8$\pm$9.2 & 18.2$\pm$10.5 & 73.5$\pm$4.6 \\
			CL-LoRA~\cite{cl_lora} & 66.8$\pm$2.6 & 63.8$\pm$2.8 & 26.8$\pm$3.5 & 79.8$\pm$7.8 & 78.6$\pm$9.5 & 19.5$\pm$11.2 & 72.8$\pm$4.8 \\
			DUCT~\cite{duct} & 67.5$\pm$2.2 & 64.2$\pm$2.5 & 24.9$\pm$2.9 & 81.2$\pm$7.2 & 80.0$\pm$9.0 & 17.6$\pm$10.8 & 74.1$\pm$4.5 \\
			\textbf{MiCo (Ours)}  & \textbf{69.5$\pm$3.2}  & \textbf{67.3$\pm$3.5}  & 23.4$\pm$5.4  & \textbf{84.6$\pm$4.1}  & \textbf{83.9$\pm$6.2}  & \textbf{11.5$\pm$4.3}  & \textbf{77.1$\pm$3.6} \\
			
			\specialrule{.05em}{.1em}{.1em} 
			& & &\multicolumn{2}{c}{\textbf{CORe50}} & & & & \\ 
			\specialrule{.05em}{.1em}{.1em} 
			\text{Fine-tuning}  & 69.6$\pm$0.3  & 69.6$\pm$0.4  & \underline{5.9$\pm$0.8}  & 45.1$\pm$4.0  & 46.4$\pm$4.4   &16.8 $\pm$0.2   & 57.4$\pm$2.2\\
			\specialrule{.05em}{.1em}{.1em} 
			EWC \cite{ewc}  & 73.0$\pm$1.8  & 67.1$\pm$2.0  & 7.5$\pm$1.5  & 48.0$\pm$2.2  & 44.2$\pm$2.5  & 19.0$\pm$3.5  & 60.7$\pm$2.0 \\
			LwF \cite{lwf}  & 73.3$\pm$1.7  & 67.4$\pm$1.9  & 7.0$\pm$1.4  & 48.1$\pm$2.1  & 44.2$\pm$2.4  & 18.0$\pm$3.2  & 60.9$\pm$1.9 \\
			L2P \cite{l2p}  & 75.5$\pm$3.5  & 75.6$\pm$3.5  & 8.0$\pm$1.9  & 50.0$\pm$1.6  & 51.6$\pm$0.9  & 18.6$\pm$2.9  & 62.8$\pm$2.6 \\
			DualPrompt \cite{dualprompt}  & 73.1$\pm$7.6  & 73.1$\pm$7.6  & 9.8$\pm$4.3  & 51.4$\pm$1.6  & 52.6$\pm$1.0  & 18.1$\pm$3.0  & 62.3$\pm$4.6\\
			CODA-Prompt \cite{codaprompt}  & 73.2$\pm$0.1  & 73.3$\pm$0.2  & 8.0$\pm$2.0  & 53.9$\pm$2.3  & 54.2$\pm$1.5  & \textbf{15.7$\pm$5.4}  & 63.6$\pm$1.2 \\
			DGR \cite{dgr}  & 77.5$\pm$3.3  & 77.0$\pm$2.9  & 12.4$\pm$3.8  & 55.8$\pm$3.6  & 56.7$\pm$2.4  & 20.1$\pm$3.5  & 66.7$\pm$3.5\\
			LAE \cite{lae}  & 72.0$\pm$3.2  & 73.2$\pm$2.6  & 8.3$\pm$2.7  & 54.1$\pm$5.0  & 55.5$\pm$5.4  & 16.6$\pm$1.0  & 63.1$\pm$4.1\\
			ICON \cite{vil}  & \underline{80.6$\pm$1.1}  & \underline{80.7$\pm$1.0}  & 6.2$\pm$0.1  & \underline{57.5$\pm$1.2}  & \underline{59.1$\pm$1.1}  & \underline{15.8$\pm$4.4}  & \underline{69.1$\pm$1.2} \\
			MISA~\cite{misa}  & 80.1$\pm$1.3  & 80.0$\pm$1.2  & 6.5$\pm$0.2  & 57.0$\pm$1.4  & 58.5$\pm$1.3  & 16.2$\pm$4.1  & 68.5$\pm$1.3 \\
			PSR~\cite{psr} & 79.8$\pm$1.3 & 79.5$\pm$1.2 & 7.5$\pm$0.8 & 56.2$\pm$1.5 & 58.2$\pm$1.4 & 18.5$\pm$4.8 & 68.5$\pm$1.4 \\
			CL-LoRA~\cite{cl_lora} & 79.2$\pm$1.5 & 78.8$\pm$1.4 & 8.1$\pm$1.0 & 55.5$\pm$1.6 & 57.5$\pm$1.5 & 19.2$\pm$5.2 & 67.8$\pm$1.5 \\
			DUCT~\cite{duct} & 80.1$\pm$1.2 & 80.2$\pm$1.1 & 6.9$\pm$0.6 & 56.8$\pm$1.4 & 58.6$\pm$1.3 & 17.8$\pm$4.6 & 68.8$\pm$1.3 \\
			\textbf{MiCo (Ours)}  & \textbf{83.5$\pm$0.8}  & \textbf{82.0$\pm$0.8 } & \textbf{5.3$\pm$0.5}  & \textbf{67.4$\pm$2.3}  & \textbf{66.4$\pm$2.0}  & 16.4$\pm$3.2  & \textbf{75.5$\pm$1.5} \\
			
			\specialrule{.05em}{.1em}{.1em} 
			& & &\multicolumn{2}{c}{\textbf{DomainNet}} & & & & \\ 
			\specialrule{.05em}{.1em}{.1em} 
			\text{Fine-tuning}  & 42.0$\pm$2.8  & 43.4$\pm$1.5  & \underline{15.2$\pm$1.9}  & 42.1$\pm$3.6  & 45.9$\pm$0.2  & \textbf{8.3$\pm$2.8}  & 42.1$\pm$3.2\\
			\specialrule{.05em}{.1em}{.1em} 
			EWC \cite{ewc}  & 44.9$\pm$2.5  & 41.3$\pm$2.8  & 21.0$\pm$3.0  & 42.7$\pm$2.6  & 39.3$\pm$2.9  & 18.0$\pm$4.0  & 43.8$\pm$2.5 \\
			LwF \cite{lwf}  & 45.1$\pm$2.4  & 41.5$\pm$2.7  & 20.0$\pm$2.8  & 42.8$\pm$2.5  & 39.4$\pm$2.8  & 16.0$\pm$3.8  & 43.8$\pm$2.4 \\
			L2P \cite{l2p}  & 47.0$\pm$1.7  & 47.3$\pm$1.2  & 24.2$\pm$1.7  & 43.2$\pm$0.9  & 41.6$\pm$0.7  & 22.4$\pm$4.9  & 45.1$\pm$1.3\\
			DualPrompt \cite{dualprompt}  & 45.8$\pm$2.9  & 47.1$\pm$1.2  & 18.6$\pm$2.7  & 46.3$\pm$3.4  & 42.0$\pm$1.9  & 23.5$\pm$5.0  & 46.1$\pm$3.2\\
			CODA-Prompt \cite{codaprompt}  & 47.6$\pm$1.9  & 48.4$\pm$0.8  & 19.9$\pm$1.7  & 46.0$\pm$2.4  & 43.2$\pm$1.9  & 24.2$\pm$3.5  & 46.8$\pm$2.2\\
			DGR \cite{dgr}  & 48.3$\pm$1.5  & 48.8$\pm$1.4  & 19.6$\pm$2.2  & 49.2$\pm$3.0  & 50.1$\pm$1.8  & 23.4$\pm$3.8  & 48.8$\pm$2.3\\
			LAE \cite{lae}  & 44.6$\pm$0.5  & 43.8$\pm$1.2  & 25.0$\pm$2.3  & 47.5$\pm$3.1  & 42.1$\pm$1.9  & 19.3$\pm$4.9  & 46.1$\pm$1.8\\
			ICON \cite{vil}  & \underline{49.8$\pm$0.3}  & \underline{52.6$\pm$1.7}  & \textbf{14.4$\pm$1.3}  & \underline{52.3$\pm$2.6}  & \underline{53.0$\pm$1.5}  & 15.2$\pm$2.6  & \underline{51.1$\pm$1.5} \\
			MISA~\cite{misa}  & 49.2$\pm$0.4  & 51.9$\pm$1.9  & 15.2$\pm$1.4  & 51.8$\pm$2.7  & 52.5$\pm$1.6  & 15.6$\pm$2.5  & 50.5$\pm$1.6 \\
			PSR~\cite{psr} & 48.5$\pm$0.5 & 51.8$\pm$1.8 & 18.5$\pm$2.5 & 51.5$\pm$2.8 & 52.2$\pm$1.6 & 17.5$\pm$2.8 & 50.5$\pm$1.6 \\
			CL-LoRA~\cite{cl_lora} & 47.9$\pm$0.6 & 51.2$\pm$1.9 & 19.8$\pm$2.8 & 50.8$\pm$2.9 & 51.5$\pm$1.7 & 18.6$\pm$3.1 & 49.8$\pm$1.7 \\
			DUCT~\cite{duct} & 49.2$\pm$0.4 & 52.1$\pm$1.6 & 17.6$\pm$2.2 & 51.9$\pm$2.5 & 52.6$\pm$1.5 & 16.8$\pm$2.5 & 50.8$\pm$1.5 \\
			\textbf{MiCo (Ours)}  & \textbf{52.3$\pm$1.2}  & \textbf{56.7$\pm$1.3}  & 16.7$\pm$1.5  & \textbf{55.2$\pm$2.3}  & \textbf{56.1$\pm$0.1}  & \underline{15.0$\pm$2.8}  & \textbf{53.8$\pm$1.8} \\
			\specialrule{.1em}{.1em}{.1em} 
		\end{tabular}
	}
	\label{tab:main}
\end{table*}

\subsection{Experiment Setup}
\label{subsec:exp_setup}

\noindent\textbf{Datasets and Metrics.} Following established protocols~\cite{vil,cofima}, our experiments are conducted on four datasets with diverse class and domain distributions: iDigits~\cite{iDigits}, CORe50~\cite{CORe50}, DomainNet~\cite{DomainNet}, and ImageNet-R~\cite{imagenet-r}, with detailed task configurations summarized in Table~\ref{tab:task_configuration}.

As per standard evaluation protocols in continual learning~\cite{l2p,codaprompt,dualprompt,vil}, we report performance after training on the final task $T$ using three metrics: \textit{Average Accuracy} (AVA), \textit{Weighted Accuracy} (WA), and \textit{Forgetting Rate} ($F$). AVA computes the unweighted mean accuracy across all tasks, while WA accounts for the imbalanced class distribution inherent to UIL by weighting each task's accuracy proportional to its sample size. $F$ measures the average performance degradation on previously learned tasks.

\noindent\textbf{Comparison Methods and Implementation Details.} We compare MiCo against state-of-the-art IL methods across five categories: (1) regularization-based methods including EWC~\cite{ewc}, LwF~\cite{lwf}, DGR~\cite{dgr}, and ICON~\cite{vil}; (2) prompt-based methods including L2P~\cite{l2p}, DualPrompt~\cite{dualprompt}, and CODA-Prompt~\cite{codaprompt}; (3) mixture-based methods (LAE~\cite{lae}); (4) recent advanced methods including PSR~\cite{psr}, CL-LoRA~\cite{cl_lora}, and DUCT~\cite{duct}; and (5) a vanilla fine-tuning baseline trained via linear probing~\cite{clip,moco,mae}. All methods uniformly adopt ViT-B/16~\cite{vit} as the frozen pre-trained backbone. Results are averaged over three random seeds. To ensure a fair comparison, we strictly follow the training configurations (learning rate, epochs, and learnable adapter setup) of ICON~\cite{vil}. For the representative buffer-based method DGR~\cite{dgr}, we maintain a fixed memory buffer with 10 samples per class for experience replay.
As illustrated in Fig.~\ref{fig:overview}, a cumulative known-class binary mask is applied to the logits during both training and inference. During training on task $t$, classes not present in the current task $\mathcal{C}_t$ are masked by setting their logits to $-\infty$, ensuring the loss is computed only over valid classes. During inference on the cumulative test set $\mathcal{D}_{test} = \{\mathcal{D}_1, \ldots, \mathcal{D}_t\}$, the mask is constructed from the union of all learned classes $\bigcup_{i=1}^{t} \mathcal{C}_i$, and the model predicts over this cumulative class set. The cumulative class set is available because UIL assumes explicit task boundaries and labeled incremental classes, consistent with standard class/domain incremental protocols. This masking solely prevents untrained future-class logits from generating meaningless predictions. This standard evaluation protocol does not require oracle task identity and is consistently applied to all compared methods to ensure a fair comparison.

\begin{table*}[]
	\centering
	\caption{Comparison with state-of-the-art IL methods on ImageNet-R dataset under UIL and VIL scenarios. Results (\%) are reported as mean $\pm$ standard deviation across three random seeds. Best and second-best results are marked in \textbf{bold} and \underline{underline}, respectively.}
	\resizebox{0.85\textwidth}{!}{
		\begin{tabular}{ccccccccc@{}}
			\specialrule{.1em}{.1em}{.1em} 
			\multirow[b]{1.8}{*}{\textbf{Method}}  & \multicolumn{3}{c}{\textbf{Versatile-IL (VIL)}}  & \multicolumn{3}{c}{\textbf{Universal-IL (UIL)}}  & \multirow[b]{1.8}{*}{\textbf{Mean Acc}($\uparrow$)}\\ 
			& AVA($\uparrow$)  & WA($\uparrow$)  & F($\downarrow$)  & AVA($\uparrow$)  & WA($\uparrow$)  & F($\downarrow$) \\
			\specialrule{.05em}{.1em}{.1em} 
			\text{Fine-tuning}  & 38.2$\pm$1.2  & 36.5$\pm$1.5  & 22.3$\pm$4.2  & 42.1$\pm$1.8  & 44.2$\pm$2.1  & 19.5$\pm$8.3  & 40.2$\pm$1.5 \\
			\specialrule{.05em}{.1em}{.1em} 
			EWC \cite{ewc}  & 40.5$\pm$2.1  & 38.2$\pm$2.3  & 26.1$\pm$3.8  & 44.5$\pm$2.5  & 43.1$\pm$2.8  & 21.2$\pm$5.1  & 42.5$\pm$2.3 \\
			LwF \cite{lwf}  & 41.2$\pm$2.0  & 39.1$\pm$2.2  & 24.5$\pm$3.5  & \underline{46.4$\pm$2.4}  & 44.5$\pm$2.6  & 20.1$\pm$4.8  & 43.8$\pm$2.2 \\
			L2P \cite{l2p}  & 43.8$\pm$3.1  & 41.5$\pm$3.4  & 27.2$\pm$4.2  & 44.8$\pm$2.3  & 44.1$\pm$2.6  & 19.5$\pm$6.5  & 44.3$\pm$2.7 \\
			DGR \cite{dgr}  & 42.1$\pm$2.6  & 40.8$\pm$2.4  & 28.5$\pm$3.9  & 44.3$\pm$3.0  & 43.5$\pm$3.3  & 22.1$\pm$4.5  & 43.2$\pm$2.8 \\
			DualPrompt \cite{dualprompt}  & 44.6$\pm$2.0  & 42.8$\pm$2.5  & 23.8$\pm$3.6  & 45.2$\pm$1.9  & 45.1$\pm$2.2  & \underline{18.2$\pm$5.1}  & 44.9$\pm$2.0 \\
			CODA-Prompt \cite{codaprompt}  & 46.8$\pm$2.2  & 44.2$\pm$2.6  & 21.5$\pm$2.4  & 44.2$\pm$3.2  & 43.8$\pm$3.0  & 18.9$\pm$6.1  & 45.5$\pm$2.7 \\
			LAE \cite{lae}  & 41.5$\pm$2.9  & 39.8$\pm$3.2  & 26.8$\pm$2.6  & 44.1$\pm$2.5  & 42.3$\pm$2.8  & 20.5$\pm$3.0  & 42.8$\pm$2.7 \\
			ICON \cite{vil}  & \underline{50.4$\pm$1.9}  & \underline{50.5$\pm$2.4}  & \underline{19.1$\pm$5.6}  & 41.3$\pm$7.9  & \underline{50.7$\pm$3.9}  & 20.5$\pm$8.5  & \underline{45.9$\pm$4.9} \\
			MISA~\cite{misa}  & 48.8$\pm$2.0  & 49.0$\pm$2.2  & 20.5$\pm$5.8  & 40.0$\pm$7.4  & 49.2$\pm$3.6  & 21.8$\pm$8.2  & 44.8$\pm$4.7 \\
			PSR~\cite{psr} & 49.2$\pm$2.1 & 49.5$\pm$2.5 & 21.5$\pm$6.2 & 40.2$\pm$7.5 & 49.5$\pm$3.8 & 23.2$\pm$8.8 & 45.2$\pm$5.1 \\
			CL-LoRA~\cite{cl_lora} & 48.5$\pm$2.3 & 48.8$\pm$2.6 & 22.8$\pm$6.8 & 39.5$\pm$7.8 & 48.8$\pm$4.0 & 24.5$\pm$9.2 & 44.5$\pm$5.3 \\
			DUCT~\cite{duct} & 49.8$\pm$2.0 & 50.1$\pm$2.3 & 20.6$\pm$5.9 & 40.8$\pm$7.2 & 50.2$\pm$3.6 & 22.1$\pm$8.5 & 45.5$\pm$4.8 \\
			\textbf{MiCo (Ours)}  & \textbf{59.3$\pm$0.9}  & \textbf{59.7$\pm$1.1}  & \textbf{11.4$\pm$1.7}  & \textbf{49.8$\pm$1.0}  & \textbf{54.4$\pm$1.4}  & \textbf{11.5$\pm$1.2}  & \textbf{54.6$\pm$0.9} \\
			\specialrule{.1em}{.1em}{.1em} 
		\end{tabular}
	}
	\label{tab:imagenet-r}
\end{table*}

\begin{table*}[!h]
	\centering
	\caption{Ablation experiments of each component in UIL. MOBJ: multi-objective learning scheme; DIR: direction recalibration module; MAG: magnitude recalibration module. We progressively integrate each component from baseline to full MiCo. DIR operates in conjunction with MOBJ to reduce conflicting gradients. Results (\%) are reported as mean $\pm$ standard deviation across three random seeds.}
	\resizebox{0.9\textwidth}{!}{% 
		
		\begin{tabular}{cccccccccccc@{}}
			\specialrule{.1em}{.1em}{.1em}
			\multicolumn{3}{c}{Method} & \multicolumn{3}{c}{\multirow{1}{*}{iDigits}} & \multicolumn{3}{c}{\multirow{1}{*}{CORe50}} & \multicolumn{3}{c}{\multirow{1}{*}{DomainNet}} \\ 
			\cmidrule(lr){1-3}\cmidrule(lr){4-6} \cmidrule(lr){7-9} \cmidrule(lr){10-12} 
			
			\multicolumn{1}{c}{\text{MOBJ}}& \multicolumn{1}{c}{\text{DIR}} & \multicolumn{1}{c}{\text{MAG}} & \multicolumn{1}{c}{AVA($\uparrow$)} & \multicolumn{1}{c}{WA($\uparrow$)} & \multicolumn{1}{c}{F($\downarrow$)} & \multicolumn{1}{c}{AVA($\uparrow$)}& \multicolumn{1}{c}{WA($\uparrow$)} & \multicolumn{1}{c}{F($\downarrow$)} &\multicolumn{1}{c}{AVA($\uparrow$)}& \multicolumn{1}{c}{WA($\uparrow$)} & \multicolumn{1}{c}{F($\downarrow$)} \\
			
			\specialrule{.05em}{.1em}{.1em} 
			&  &  & 78.3$\pm$3.2 & 79.5$\pm$6.1 & 14.3$\pm$7.9  & 64.9$\pm$5.6 & 64.7$\pm$6.9  & 20.5$\pm$ 4.3  & 46.9$\pm$3.8 & 55.2$\pm$1.4 & 11.6$\pm$2.6 \\
			\checkmark &  &  & 80.4$\pm$4.6 & 81.0$\pm$5.9 & 13.5$\pm$9.5 & 65.0$\pm$3.2 & 65.5$\pm$4.4 & 19.7$\pm$2.3 & 49.8$\pm$3.0 & 55.7$\pm$1.4 & 16.6$\pm$1.4 \\
			&  &\checkmark & 81.3$\pm$4.4 & 82.4$\pm$6.3 & 13.0$\pm$9.1 & 65.9$\pm$3.8 & 65.9$\pm$4.6 & 18.3$\pm$4.2 & 50.2$\pm$3.1 & 54.5$\pm$1.4 & \textbf{14.9$\pm$1.9} \\
			\checkmark&  &\checkmark & 84.0$\pm$3.0 & 83.9$\pm$4.7 & 12.4$\pm$6.9 & 67.1$\pm$2.5 & \textbf{66.9$\pm$2.1} & 17.1$\pm$2.8 &53.1$\pm$2.6 & 55.9$\pm$0.3 & 14.5$\pm$0.7 \\
			\checkmark&\checkmark&  & 83.1$\pm$3.9 & 82.5$\pm$6.0 & 13.4$\pm$7.4 & 66.4$\pm$1.2 & 67.3$\pm$2.3 & 19.0$\pm$2.3 & 51.2$\pm$3.1 & 55.2$\pm$0.3 & 15.8$\pm$5.1 \\
			\checkmark&\checkmark &\checkmark  & \textbf{84.6$\pm$4.1} & \textbf{83.9$\pm$6.2 }& \textbf{11.5$\pm$4.3} & \textbf{67.4$\pm$2.3} & 66.4$\pm$2.0 & \textbf{16.4$\pm$3.2} & \textbf{55.2$\pm$2.3} & \textbf{56.1$\pm$0.1} & 15.0$\pm$2.8 \\
			\specialrule{.1em}{.1em}{.1em} 
		\end{tabular}
		
	}
	
	\label{tab:ablation}
\end{table*}

\subsection{Comparison with State-of-the-Art IL Methods}
\label{subsec:main_results}

We evaluate MiCo against existing methods under both the proposed UIL and the existing VIL scenarios. As shown in Table~\ref{tab:main}, MiCo consistently outperforms all baselines across every evaluation metric. Compared to the second-best results in terms of Mean Acc ($\uparrow$) across the three benchmarks (74.9, 69.1, and 51.1), MiCo achieves substantial gains of \textbf{77.1 (+2.2)}, \textbf{75.5 (+6.4)}, and \textbf{53.8 (+2.7)}, while maintaining competitive or lower variance (\textbf{3.6}, \textbf{1.5}, \textbf{1.8}). Furthermore, MiCo achieves a remarkably low forgetting rate $F$, demonstrating its robust capacity to mitigate both inter- and intra-task distribution confusion without compromising historical knowledge retention.

We analyze the limitations of existing paradigms under UIL:
\begin{itemize}
	\item \textit{Regularization-based methods} (EWC~\cite{ewc}, LwF~\cite{lwf}, ICON~\cite{vil}, DGR~\cite{dgr}) show limited adaptability to random increment types and scales. ICON's orthogonality constraint on adapter shifts degrades under random scales, and DGR's rehearsal-based distillation becomes unreliable when types vary arbitrarily.
	\item \textit{Prompt-based methods} (L2P~\cite{l2p}, DualPrompt~\cite{dualprompt}, CODA-Prompt~\cite{codaprompt}) rely on cosine-similarity prompt selection, which is highly sensitive to unpredictable domain shifts within classes.
	\item \textit{Mixture-based methods} (LAE~\cite{lae}) continually merge online and offline models but show considerable performance fluctuations in highly confused IL settings.
	\item \textit{Recent advanced methods} (PSR~\cite{psr}, CL-LoRA~\cite{cl_lora}, DUCT~\cite{duct}) excel in their specific single-type or static-imbalance settings but lack explicit decoupled mechanisms for the dual randomness of UIL.
\end{itemize}

Furthermore, MISA~\cite{misa}, a replay-independent GCL method, while effective under controlled settings, its static optimization strategy lacks explicit mechanisms for UIL's dual randomness, yielding suboptimal performance compared to our dynamic gradient recalibration approach.

To further validate MiCo's generalization under complex appearance variations and domain shifts, we extend our evaluation to ImageNet-R (Table~\ref{tab:imagenet-r}). MiCo achieves a mean accuracy of \textbf{54.6}, substantially outperforming the second-best ICON (45.9) and MISA (44.8), while maintaining exceptionally low forgetting rates in both VIL ($11.4$) and UIL ($11.5$). These results confirm that MiCo's confusion-mitigation strategy is robust and generalizable across diverse benchmarks.

Despite natural class overlap in UIL aiding even vanilla fine-tuning, MiCo significantly surpasses this baseline, validating the necessity of our targeted confusion-mitigation strategy.

\begin{figure}[]
	\centering
	\includegraphics[width=\columnwidth]{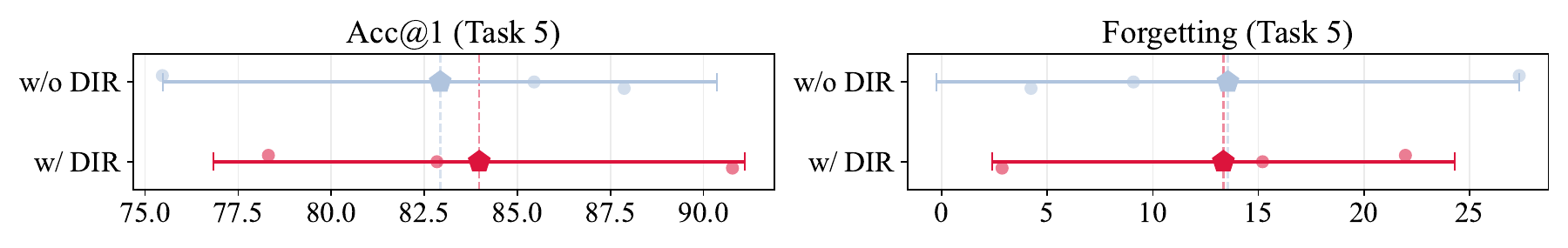}
	\caption{Effect of the direction recalibration (DIR) module on the final task of iDigits (5-task split). DIR consistently improves Acc@1 and reduces forgetting, confirming its role in stabilizing multi-objective optimization.}
	\label{fig:dir_task5}
\end{figure}

\begin{table}[]
	\centering
	\caption{Head/mid/tail analysis on ImageNet-R under the UIL setting. Classes are grouped by frequency into head (top 40\%), mid (middle 40\%), and tail (bottom 20\%) groups. MAG consistently improves all groups, with the most significant gain on tail classes. Overall accuracy (\%) is computed on all test samples, not as the unweighted average of group accuracies.}
	\resizebox{\columnwidth}{!}{
		\begin{tabular}{lcccc}
			\specialrule{.1em}{.1em}{.1em}
			Method & Head Acc & Mid Acc & Tail Acc & Overall Acc \\
			\specialrule{.05em}{.1em}{.1em}
			Baseline & 46.0$\pm$1.6 & 40.0$\pm$1.8 & 31.0$\pm$2.0 & 42.1$\pm$1.8 \\
			MAG & 47.2$\pm$1.5 & 42.4$\pm$1.6 & 39.3$\pm$1.9 & 44.9$\pm$1.9 \\
			MiCo & \textbf{49.6$\pm$1.2} & \textbf{45.1$\pm$1.3} & \textbf{41.8$\pm$1.6} & \textbf{49.8$\pm$1.0} \\
			\specialrule{.1em}{.1em}{.1em}
		\end{tabular}
	}
	\label{tab:mag_head_mid_tail_imagenetr}
\end{table}

\subsection{Ablation Study}
\label{subsec:ablation_study}

\noindent\textbf{Effectiveness of Components.} Table~\ref{tab:ablation} presents systematic ablation by progressively integrating each component into our baseline. From a Mean Acc of $63.3\!\pm\!4.2$, introducing the multi-objective learning scheme (MOBJ) yields a gain of $1.8$ points ($65.1\!\pm\!3.6$), validating its effectiveness in enforcing deterministic predictions under random increment types. Independently adding the magnitude recalibration module (MAG) improves performance by $2.5$ points ($65.8\!\pm\!3.7$), confirming its ability to alleviate asymmetric optimization caused by imbalanced class distributions. When MOBJ and MAG are combined, their synergy produces an absolute gain of $3.0$ points. To isolate the effect of the direction recalibration module (DIR), we enable it alongside MOBJ, yielding consistent improvements across all benchmarks. Overall, the full MiCo framework outperforms the baseline by up to \textbf{5.8} points in accuracy and reduces the forgetting rate by up to \textbf{1.2} points, confirming that each component is both necessary and complementary.

\noindent\textbf{Analysis of DIR.} To visualize DIR's impact, we conduct a case study on iDigits using a 5-task split. As shown in Fig.~\ref{fig:dir_task5}, the variant equipped with DIR consistently achieves higher Acc@1 and lower forgetting on the final task compared to the ablated version, validating DIR's capacity to mitigate gradient conflicts and stabilize optimization.

\begin{figure}[]
	\centering
	\includegraphics[width=0.9\columnwidth]{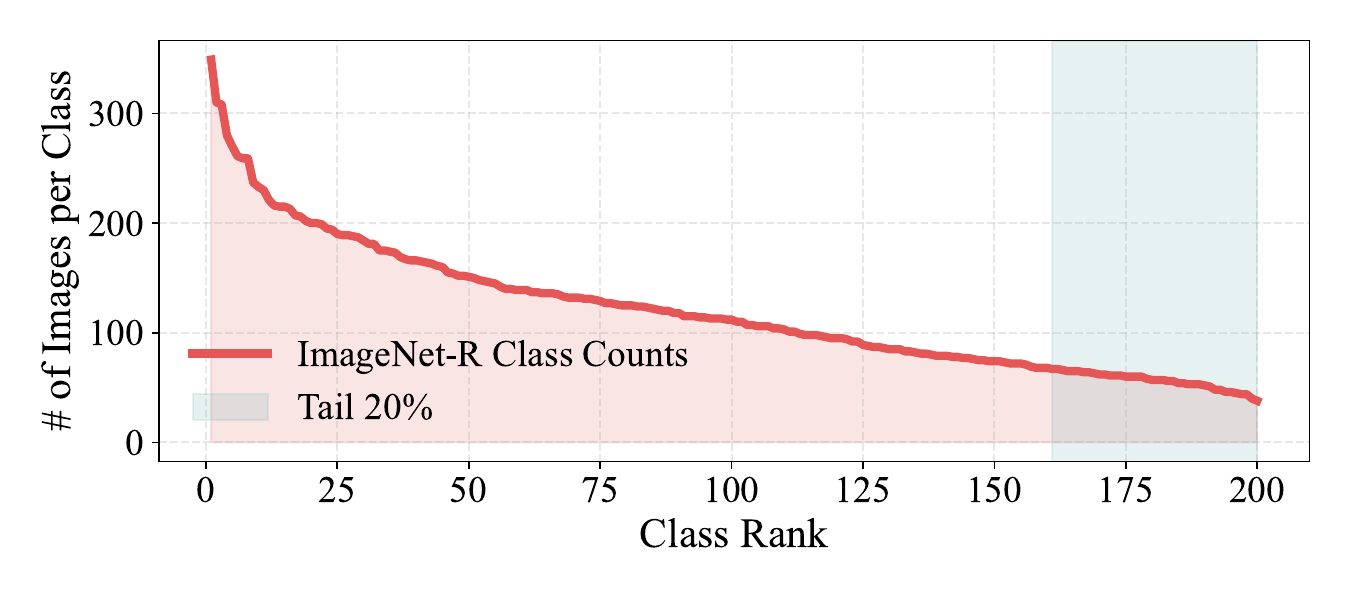}
	\caption{Class-frequency distribution of ImageNet-R under the UIL split. Classes are sorted in descending order by training sample count, revealing a pronounced long-tail distribution.}
	\label{fig:imagenetr_longtail}
\end{figure}

\noindent\textbf{Long-tail Distribution and MAG on ImageNet-R.} Fig.~\ref{fig:imagenetr_longtail} illustrates that ImageNet-R exhibits a severe long-tail distribution under the UIL protocol, with the tail 20\% of classes receiving significantly fewer training samples. As shown in Table~\ref{tab:mag_head_mid_tail_imagenetr}, MAG consistently improves accuracy across all groups, with the most substantial gain observed on the tail classes. When integrated into the full MiCo framework, these gains are further amplified while maintaining low variance. \textit{Note:} Overall accuracy is computed over all test samples, not as an unweighted average of group accuracies.

\begin{table}[]
	\centering
	\caption{Comparison with imbalance-aware losses on iDigits under the UIL setting. Results (\%) are reported as mean $\pm$ standard deviation over three random seeds. Lower is better for Train Loss, Test Loss, and Forgetting; higher is better for AVA and WA. MiCo achieves the lowest test loss and best overall performance, demonstrating superior generalization and knowledge retention.}
	\resizebox{\columnwidth}{!}{
		\begin{tabular}{lccccc}
			\specialrule{.1em}{.1em}{.1em}
			Method & Train Loss$\downarrow$ & Test Loss$\downarrow$ & AVA$\uparrow$ & WA$\uparrow$ & F$\downarrow$ \\
			\specialrule{.05em}{.1em}{.1em}
			CE & 0.137$\pm$0.076 & 1.687$\pm$0.089 & 81.1$\pm$8.6 & 82.5$\pm$6.5 & 17.0$\pm$11.4 \\
			Focal & \textbf{0.007$\pm$0.006} & 1.987$\pm$0.041 & 76.0$\pm$7.6 & 79.3$\pm$4.0 & 17.4$\pm$9.1 \\
			LDAM & 2.067$\pm$1.018 & 2.007$\pm$0.048 & 77.5$\pm$10.2 & 79.8$\pm$6.9 & 20.9$\pm$12.7 \\
			MiCo & 0.160$\pm$0.088 & \textbf{1.654$\pm$0.054} & \textbf{84.6$\pm$4.1} & \textbf{83.9$\pm$6.2} & \textbf{11.5$\pm$4.3} \\
			\specialrule{.1em}{.1em}{.1em}
		\end{tabular}
	}
	\label{tab:imbalance_loss}
\end{table}

\noindent\textbf{Comparison with Imbalance-Aware Losses.} We compare MiCo against established imbalance-aware losses, including Focal Loss~\cite{focal} and LDAM~\cite{ldam}, on iDigits under UIL. As shown in Table~\ref{tab:imbalance_loss}, MiCo achieves the lowest test loss and the highest AVA and WA, with notably reduced forgetting. Since Focal Loss and LDAM target \textit{static} class imbalance, they cannot handle UIL's \textit{dynamic} inter-task distribution shifts; MiCo's gradient recalibration addresses both direction and magnitude dynamically.

\begin{table}[]
	\centering
	\caption{Performance comparison under the Strictly Disjoint UIL setting on DomainNet.}
	\label{tab:strictly_disjoint}
	\resizebox{\columnwidth}{!}{
		\begin{tabular}{lcccc}
			\toprule
			\textbf{Method} & \textbf{AVA ($\uparrow$)} & \textbf{WA ($\uparrow$)} & \textbf{Forgetting ($\downarrow$)} & \textbf{$\Delta$ vs. Fine-tuning} \\
			\midrule
			Vanilla Fine-tuning & 32.4 $\pm$ 1.8 & 34.1 $\pm$ 2.1 & 28.6 $\pm$ 3.5 & -- \\
			ICON    & 39.7 $\pm$ 1.5 & 41.2 $\pm$ 1.8 & 22.4 $\pm$ 2.8 & +7.3 \\
			PSR     & 38.5 $\pm$ 1.6 & 39.8 $\pm$ 1.9 & 24.2 $\pm$ 3.1 & +6.1 \\
			CL-LoRA & 39.4 $\pm$ 1.4 & 40.7 $\pm$ 1.7 & 23.5 $\pm$ 2.8 & +7.0 \\
			DUCT    & 37.8 $\pm$ 1.7 & 39.2 $\pm$ 2.0 & 25.1 $\pm$ 3.2 & +5.4 \\
			MISA    & 40.2 $\pm$ 1.3 & 41.8 $\pm$ 1.6 & 21.9 $\pm$ 2.5 & +7.8 \\
			\textbf{MiCo (Ours)} & \textbf{46.8 $\pm$ 1.1} & \textbf{48.5 $\pm$ 1.3} & \textbf{15.2 $\pm$ 2.1} & \textbf{+14.4} \\
			\bottomrule
		\end{tabular}
	}
\end{table}

\noindent\textbf{Strictly Disjoint UIL Setting.} To address concerns regarding the implicit replay effect caused by natural class overlap in standard UIL, we construct a stringent \textit{Strictly Disjoint UIL} protocol on DomainNet, where each class appears exactly once across all tasks, completely eliminating inter-task data overlap. As shown in Table~\ref{tab:strictly_disjoint}, all methods suffer performance degradation compared to standard UIL, confirming that implicit replay offers only marginal regularization. Crucially, MiCo maintains an AVA of \textbf{46.8\%}, outperforming the strongest baseline MISA (40.2\%) by 6.6 percentage points. This confirms that MiCo's superiority stems from principled gradient recalibration, not incidental data redundancy.

\begin{table*}[]
\caption{Ablation of the balancing coefficient $\gamma$ in the multi-objective learning scheme. We evaluate $\gamma \in \{1.0, 0.1, 0.01, 0.001\}$ across three benchmarks. $\gamma = 0.01$ achieves the best trade-off between accuracy and forgetting, and is adopted as the default setting. Results (\%) are reported as mean $\pm$ standard deviation across three random seeds.}
\centering
    \resizebox{\textwidth}{!}{

\begin{tabular}{lccccccccc@{}}
        \specialrule{.1em}{.1em}{.1em}
        \multirow[b]{1.8}{*}{$\gamma$} & \multicolumn{3}{c}{\multirow{1}{*}{\textbf{iDigits}}} & \multicolumn{3}{c}{\multirow{1}{*}{\textbf{CORe50}}} & \multicolumn{3}{c}{\multirow{1}{*}{\textbf{DomainNet}}} \\ 
            \cmidrule(lr){2-4} \cmidrule(lr){5-7} \cmidrule(lr){8-10}
             & \multicolumn{1}{c}{AVA($\uparrow$)} & \multicolumn{1}{c}{WA($\uparrow$)} & \multicolumn{1}{c}{F($\downarrow$)} & \multicolumn{1}{c}{AVA($\uparrow$)} & \multicolumn{1}{c}{WA($\uparrow$)} & \multicolumn{1}{c}{F($\downarrow$)} &\multicolumn{1}{c}{AVA($\uparrow$)} & \multicolumn{1}{c}{WA($\uparrow$)} & \multicolumn{1}{c}{F($\downarrow$)} \\
             
             \specialrule{.05em}{.1em}{.1em} 
             1.0 &\textbf{ 85.1$\pm$4.3} &\textbf{84.0$\pm$ 5.3} & 12.7$\pm$7.9 & 66.3$\pm$3.4 & 66.6$\pm$2.8 & 17.2$\pm$2.1 & 47.3$\pm$3.9 & 54.4$\pm$0.5 & 11.1$\pm$1.7 \\
             0.1 & 80.5$\pm$7.9 &80.8$\pm$ 8.4 & 15.8$\pm$10.2 & 66.5$\pm$4.5 & 66.4$\pm$5.6 & 17.8$\pm$5.2 & 52.0$\pm$1.2 & 55.5$\pm$0.7 & 15.3$\pm$4.9 \\
             0.01 & 84.6$\pm$4.1 &83.9$\pm$6.2 & \textbf{11.5$\pm$4.3} & \textbf{67.4$\pm$2.3} & \textbf{66.4$\pm$2.0} & \textbf{16.4$\pm$3.2} & \textbf{55.2$\pm$2.3} & \textbf{56.1$\pm$0.1} &  \textbf{15.0$\pm$2.8} \\
             0.001 & 83.2$\pm$2.1 &82.1$\pm$5.8 & 13.6$\pm$7.6 & 66.5$\pm$2.1  & 66.2$\pm$0.3 & 22.0$\pm$0.6  & 52.0$\pm$0.1 & 54.9$\pm$0.1 &  19.1$\pm$1.3 \\
            
            \specialrule{.1em}{.1em}{.1em} 
    \end{tabular}

    }
\label{tab:gamma}
\end{table*}

\noindent\textbf{Ablation of $\gamma$.} We tune the balancing coefficient $\gamma \in \{1.0, 0.1, 0.01, 0.001\}$ across three random seeds. As shown in Table~\ref{tab:gamma}, $\gamma = 0.01$ achieves peak AVA on CORe50~\cite{CORe50} and DomainNet~\cite{DomainNet}. Although it does not yield the absolute highest accuracy on iDigits~\cite{iDigits}, it maintains comparable accuracy while achieving the lowest forgetting rate. Larger $\gamma$ values degrade accuracy by over-penalizing entropy, while smaller values fail to enforce deterministic predictions. Thus, we adopt $\gamma = 0.01$ as the default.

\begin{figure}[]
	\centering
	\includegraphics[width=0.5\columnwidth]{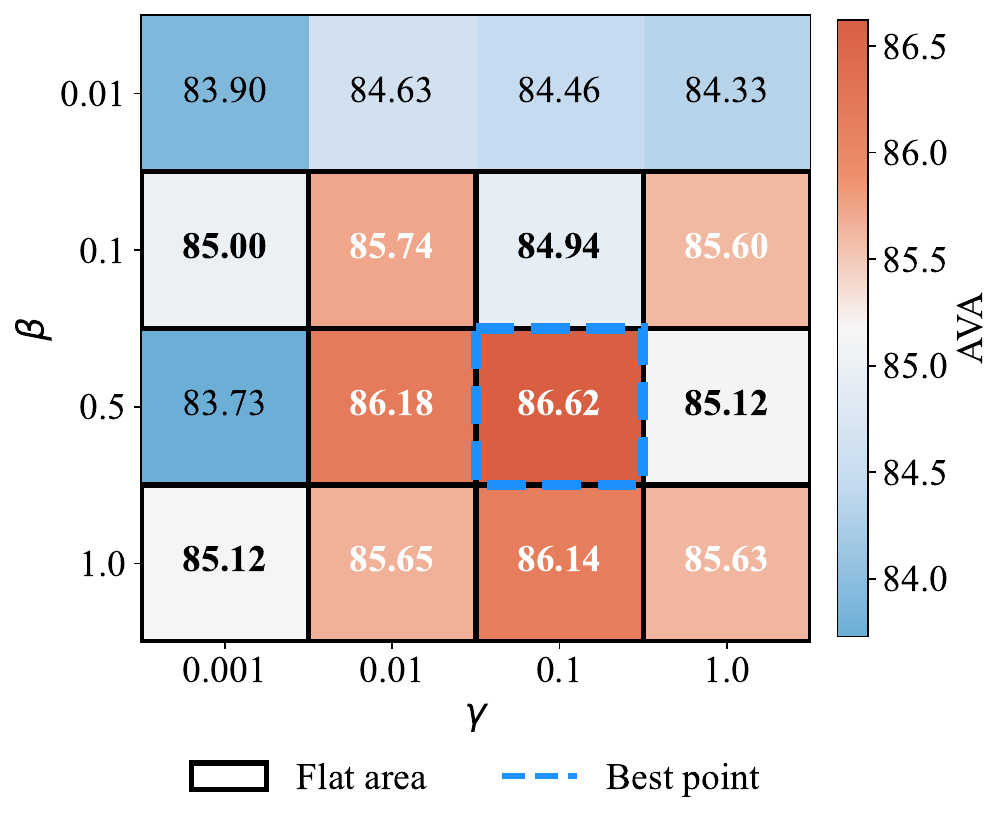}
	\caption{Sensitivity analysis of $\beta$ and $\gamma$ on iDigits under UIL. Black dashed boxes highlight flat high-performance regions, while the blue box marks the grid-search optimum.}
	\label{fig:beta_gamma_heatmap}
\end{figure}

\noindent\textbf{Sensitivity of $\beta$ and $\gamma$.} Beyond isolated $\gamma$ tuning, we examine the joint landscape of $\beta$ and $\gamma$ via a heatmap on iDigits (Fig.~\ref{fig:beta_gamma_heatmap}). Favorable configurations form a broad, flat high-performance region rather than concentrating at a single point, demonstrating that MiCo is not overly sensitive to hyperparameter choices and exhibits strong practical robustness.

\begin{table*}[]
	\centering
	\caption{Efficiency comparison on iDigits under the UIL setting with five tasks. Results are reported as mean $\pm$ standard deviation over three random seeds. Iter Time and Epoch Time denote the average training time per iteration and per epoch, respectively. Lower is better for all metrics. MiCo achieves the best efficiency across FLOPs, time, and memory while maintaining superior accuracy.}
	\resizebox{\textwidth}{!}{
		\begin{tabular}{lccccc}
			\specialrule{.1em}{.1em}{.1em}
			Method  & Train GFLOPs/iter$\downarrow$ & Iter Time (s)$\downarrow$  & Epoch Time (s)$\downarrow$  & Peak Memory (MB)$\downarrow$  & Total Time (s)$\downarrow$ \\
			\specialrule{.05em}{.1em}{.1em}
			Baseline  & 1265.9 & 0.248$\pm$0.039  & 1308.8$\pm$269.0  & 2239.7$\pm$0.6  & 42867.7$\pm$7237.5 \\
			ICON  & 1785.2 & 0.198$\pm$0.017  & 1047.5$\pm$183.1  & 3996.0$\pm$0.0  & 36046.7$\pm$2422.8 \\
			MiCo  & 1420.5 & \textbf{0.181$\pm$0.030}  & \textbf{949.6$\pm$183.0}  & \textbf{2268.0$\pm$0.0}  & \textbf{30101.7$\pm$4674.2} \\
			\specialrule{.1em}{.1em}{.1em}
		\end{tabular}
	}
	\label{tab:efficiency}
\end{table*}

\subsection{Efficiency Analysis and Scalability}
\label{subsec:efficiency}

We compare the training efficiency of MiCo against the baseline and ICON on iDigits (5-task split) using a single NVIDIA RTX 4090 GPU. As shown in Table~\ref{tab:efficiency}, MiCo achieves the best efficiency across all metrics: FLOPs, iteration time, epoch time, peak memory, and total training time. While MiCo introduces a marginal FLOPs increase over the Baseline (+12.2\%), it significantly reduces FLOPs compared to ICON (-20.4\%), as DIR and MAG operate exclusively on lightweight classifier weights. Compared to ICON, MiCo reduces iteration time by 8.7\%, epoch time by 9.3\%, total training time by 16.5\%, and peak memory by 43.2\%. Against the baseline, MiCo also achieves substantial speedups with nearly identical memory footprint, while elevating average accuracy from $78.3\!\pm\!3.2$ to $84.6\!\pm\!4.1$.

\noindent \textbf{Scalability on Massive Diverse Data.} To evaluate scalability limits, we design a series of UIL protocols on DomainNet with increasing increment scales by varying the total number of tasks ($T \in \{40, 30, 20, 10\}$). As shown in Table~\ref{tab:scalability}, as the increment scale expands, the performance advantage of MiCo consistently widens. Under the extreme scale ($T=10$), \textit{MiCo w/o DIR} and \textit{MiCo w/o MAG} degrade significantly, confirming the indispensability of decoupled recalibration at large scales.

\begin{table}[!htbp]
	\centering
	\caption{Scalability analysis on DomainNet under varying increment scales. Smaller $T$ dictates a larger number of classes/domains per task, intensifying the distribution randomness.}
	\label{tab:scalability}
	\resizebox{\columnwidth}{!}{
		\begin{tabular}{ccccc}
			\toprule
			\textbf{Scale (Classes/Task)} & \textbf{Method} & \textbf{AVA ($\uparrow$)} & \textbf{F ($\downarrow$)} & \textbf{$\Delta$ vs. ICON} \\
			\midrule
			\multirow{4}{*}{~8 ($T=40$)} & ICON & 52.3 $\pm$ 1.5 & 15.2 $\pm$ 2.6 & -- \\
			& MiCo w/o DIR & 53.1 $\pm$ 1.6 & 14.5 $\pm$ 2.4 & +0.8 \\
			& MiCo w/o MAG & 53.8 $\pm$ 1.4 & 13.9 $\pm$ 2.5 & +1.5 \\
			& \textbf{MiCo} & \textbf{55.2 $\pm$ 1.2} & \textbf{12.8 $\pm$ 2.1} & \textbf{+2.9} \\
			\midrule
			\multirow{4}{*}{~11 ($T=30$)} & ICON & 48.6 $\pm$ 1.8 & 19.4 $\pm$ 3.1 & -- \\
			& MiCo w/o DIR & 49.5 $\pm$ 1.7 & 18.2 $\pm$ 2.9 & +0.9 \\
			& MiCo w/o MAG & 50.7 $\pm$ 1.6 & 17.1 $\pm$ 2.8 & +2.1 \\
			& \textbf{MiCo} & \textbf{52.4 $\pm$ 1.4} & \textbf{14.6 $\pm$ 2.4} & \textbf{+3.8} \\
			\midrule
			\multirow{4}{*}{~17 ($T=20$)} & ICON & 44.1 $\pm$ 2.2 & 24.5 $\pm$ 3.8 & -- \\
			& MiCo w/o DIR & 45.2 $\pm$ 2.0 & 22.8 $\pm$ 3.5 & +1.1 \\
			& MiCo w/o MAG & 46.8 $\pm$ 1.9 & 21.3 $\pm$ 3.2 & +2.7 \\
			& \textbf{MiCo} & \textbf{48.9 $\pm$ 1.7} & \textbf{17.2 $\pm$ 2.9} & \textbf{+4.8} \\
			\midrule
			\multirow{4}{*}{~34 ($T=10$)} & ICON & 38.5 $\pm$ 2.8 & 31.2 $\pm$ 4.5 & -- \\
			& MiCo w/o DIR & 40.2 $\pm$ 2.6 & 28.5 $\pm$ 4.1 & +1.7 \\
			& MiCo w/o MAG & 41.5 $\pm$ 2.5 & 26.9 $\pm$ 3.9 & +3.0 \\
			& \textbf{MiCo} & \textbf{44.7 $\pm$ 2.1} & \textbf{21.4 $\pm$ 3.3} & \textbf{+6.2} \\
			\bottomrule
		\end{tabular}
	}
\end{table}

\section{Discussion on Broader Applicability and Extensions}

Beyond standard image benchmarks, UIL principles are highly relevant to complex data streams with constantly shifting distributions. For instance, MiCo's buffer-free gradient recalibration offers a robust solution for continuous learning in evolving smart city graphs~\cite{smartcity_graph} and multi-scale environmental forecasting~\cite{streamflow_cnn_transformer}, effectively addressing unpredictable spatial and temporal dynamics. Specifically, DIR resolves gradient conflicts from abrupt graph topological shifts, while MAG mitigates bias from skewed sensor frequencies. Moreover, its buffer-free nature enables deployment on resource-constrained edge devices.

Furthermore, MiCo's buffer-free optimization principles show strong potential for Multimodal Large Language Models (MLLMs). Specifically, integrating Direction Recalibration (DIR) into visual adapters effectively mitigates severe visual forgetting from unpredictable scenario shifts~\cite{mllm_mcl}, addressing inter-task confusion. Meanwhile, Magnitude Recalibration (MAG) counteracts parameter drift and long-tailed bias induced by continuous multimodal adaptation~\cite{mixture_of_noise}. Given the prohibitive memory cost of exemplar replay in massive MLLMs, our buffer-free design is particularly advantageous. Moreover, operating directly on tunable gradients enables seamless integration with existing PEFT strategies.

\section{Conclusion}

In this paper, we investigate Universal Incremental Learning (UIL), a realistic setting that removes prior assumptions on the type and scale of increments while retaining explicit task boundaries. We systematically analyze the UIL scenario and identify the dual confusion arising from inter- and intra-task distribution randomness. To mitigate these challenges, we propose MiCo, a buffer-free framework that integrates a multi-objective learning scheme with direction- and magnitude-decoupled gradient recalibration modules. Furthermore, theoretical intuition explains how MiCo mitigates representation drift through first-order surrogate-based gradient alignment and balances convergence via diagonal preconditioning. Extensive experiments, including rigorous strictly disjoint and scalability evaluations, demonstrate that MiCo achieves state-of-the-art performance across diverse benchmarks. We believe the UIL paradigm will benefit the incremental learning community by shifting evaluation from static task boundaries to dynamic distribution randomness. 

More broadly, MiCo's principles show significant potential for complex data streams and Multimodal Large Language Models. While our current ``Universal'' scope remains confined to explicit label spaces, precluding unknown category discovery, future work will establish comprehensive UIL benchmarks, extend the framework to fully task-free settings, and explore deeper theoretical convergence guarantees.

\section*{Declaration of Competing Interest}
The authors declare that they have no known competing financial interests or personal relationships that could have appeared to influence the work reported in this paper.

\section*{Acknowledgments}
This work was supported by the National Science Fund of China under Grant Nos. 62476054 and 62576153.

\printcredits
\bibliographystyle{cas-model2-names}
\bibliography{cas-refs}

\end{document}